\documentclass[10pt,twocolumn,letterpaper]{article}

\usepackage{iccv}
\usepackage{times}
\usepackage{epsfig}
\usepackage{graphicx}
\usepackage{amsmath}
\usepackage{amssymb}
\usepackage{booktabs}
\usepackage{xcolor}
\usepackage{bm}
\usepackage{mathtools}
\usepackage{multirow}
\usepackage{comment}
\usepackage{microtype}
\usepackage{authblk}
\usepackage[subtle]{savetrees}
\usepackage{pdfpages}

\newif\ifdrafting
\draftingtrue 

\ifdrafting
    \newcommand{\ec}[1]{\textcolor{cyan}{[Eric: #1]}}
    \newcommand{\jj}[1]{\textcolor{green}{[JJ: #1]}}
    \newcommand{\kn}[1]{\textcolor{red}{[Koki: #1]}}
    \newcommand{\ab}[1]{\textcolor{magenta}{[Alex: #1]}}
    
    \newcommand{\gw}[1]{\textcolor{blue}{[GW: #1]}}
    \newcommand{\tk}[1]{\textcolor{olive}{[Tero: #1]}}
    \newcommand{\sd}[1]{\textcolor{brown}{[Shalini: #1]}}
    \newcommand{\mc}[1]{\textcolor{magenta}{[Matthew: #1]}}
\else
    \newcommand{\ec}[1]{}
    \newcommand{\jj}[1]{}
    \newcommand{\kn}[1]{}
    \newcommand{\ab}[1]{}
    \newcommand{\gw}[1]{}
    \newcommand{\tk}[1]{}
    \newcommand{\mc}[1]{}
    \newcommand{\sd}[1]{}
    
\fi

\usepackage[pagebackref=true,breaklinks=true,letterpaper=true,colorlinks,bookmarks=false]{hyperref}

\usepackage[capitalize]{cleveref}
\crefname{section}{Sec.}{Secs.}
\Crefname{section}{Section}{Sections}
\Crefname{table}{Table}{Tables}
\crefname{table}{Tab.}{Tabs.}

\iccvfinalcopy 

\def\iccvPaperID{10260} 

\begin{document}

\title{Generative Novel View Synthesis with 3D-Aware Diffusion Models}


\makeatletter
    \renewcommand\AB@affilsepx{ \hphantom{---} \protect\Affilfont}
\makeatother

\newcommand*\samethanks[1][\value{footnote}]{\footnotemark[#1]}

\author[1,2]{Eric R. Chan \thanks{Equal contribution.}\thanks{Work was done during an internship at NVIDIA.}}
\author[2]{Koki Nagano\samethanks[1]}
\author[2]{Matthew A. Chan\samethanks[1]}
\author[1]{Alexander W. Bergman\samethanks[1]}
\author[1]{Jeong Joon Park\samethanks[1]}
\author[1]{\\Axel Levy}
\author[2]{Miika Aittala}
\author[2]{Shalini De Mello}
\author[2]{Tero Karras}
\author[1]{Gordon Wetzstein}
\affil[1]{Stanford University}
\affil[2]{NVIDIA}

\maketitle

\begin{abstract}
We present a diffusion-based model for 3D-aware generative novel view synthesis from as few as a single input image. Our model samples from the distribution of possible renderings consistent with the input and, even in the presence of ambiguity, is capable of rendering diverse and plausible novel views. To achieve this, our method makes use of existing 2D diffusion backbones but, crucially, incorporates geometry priors in the form of a 3D feature volume. This latent feature field captures the distribution over possible scene representations and improves our method's ability to generate view-consistent novel renderings. In addition to generating novel views, our method has the ability to autoregressively synthesize 3D-consistent sequences. We demonstrate state-of-the-art results on synthetic renderings and room-scale scenes; we also show compelling results for challenging, real-world objects.
\end{abstract}

\section{Introduction}
\label{sec:intro}

\begin{figure}[t!]
    \centering
    \includegraphics[width=\linewidth]{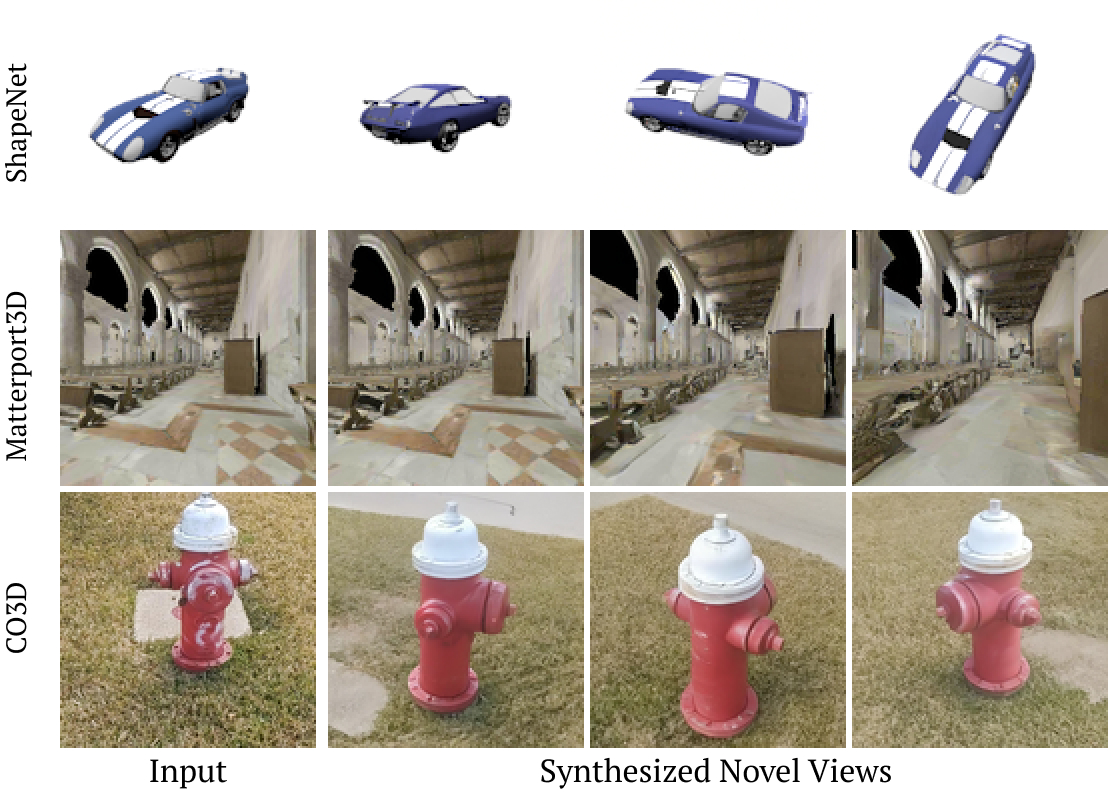}
    \caption{Our 3D-aware diffusion model synthesizes realistic novel views from as little as a single input image. These results are generated with the ShapeNet~\cite{chang2015shapenet}, Matterport3D~\cite{chang2017matterport3d}, and Common Objects in 3D~\cite{reizenstein2021common} datasets.}
    \label{fig:teaser}
    \vspace{-5pt}
\end{figure}

\begin{figure}[t!]
    \centering
    \includegraphics[width=\linewidth]{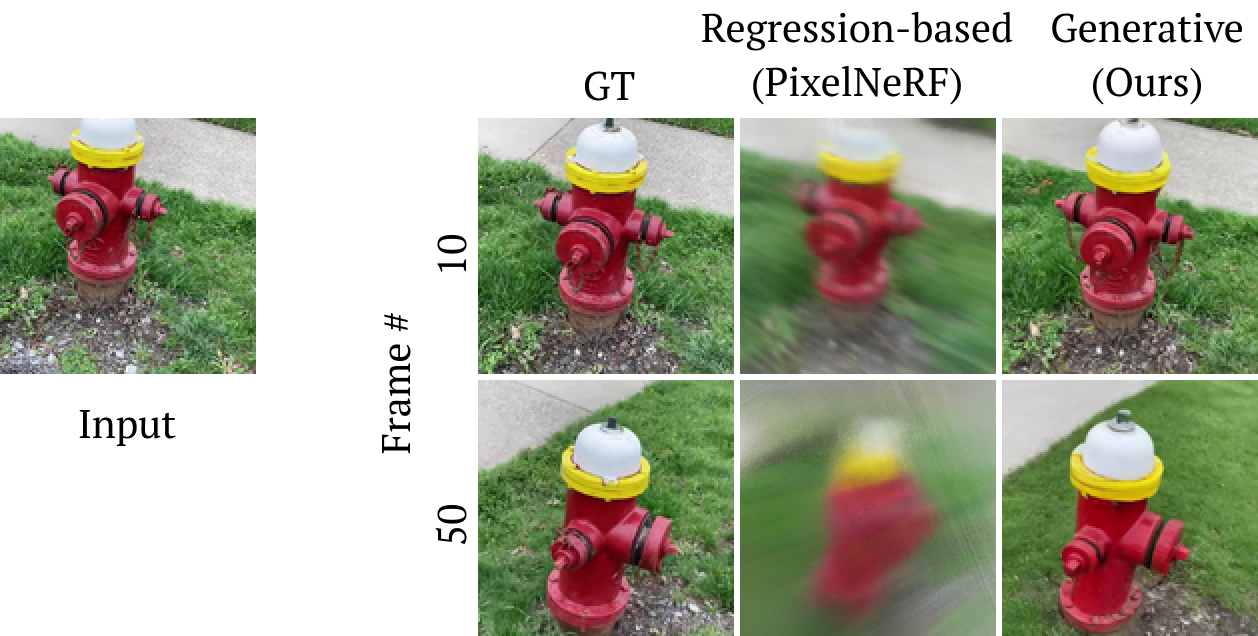}
    \caption{While regression-based models are capable of effective view synthesis near input views (top row), they blur across ambiguity when extrapolating. Generative approaches can continue to sample plausible renderings far from input views (second row, third column).}
    \label{fig:generative_motivation}
    \vspace{-5pt}
\end{figure}

In this work, we challenge ourselves to addresses multiple open problems in novel view synthesis (NVS): to design an NVS framework that (1) operates from as little as a single image and is capable of (2) generating long-range of sequences far from the input views as well as (3) handling both individual objects and complex scenes (see Fig.~\ref{fig:teaser}). While existing few-shot NVS approaches, trained on a category of objects with a regression objective, can generate geometrically consistent renderings, i.e., sequences whose frames share a coherent scene structure, they are ineffective in handling extrapolation and unbounded scenes (see Fig.~\ref{fig:generative_motivation}). Dealing with long-range extrapolation (2) requires using a generative prior to deal with the innate ambiguity that comes with completing portions of the scenes that were unobserved in the input. In this work, we propose a diffusion-based few-shot NVS framework that can generate plausible and competitively geometrically consistent renderings, pushing the boundaries of NVS towards a solution that can operate in a wide range of challenging real-world data.

Previous approaches to few-shot novel view synthesis can broadly be grouped into two categories.
Geometry-prior-based methods~\cite{Riegler2021SVS,Riegler2020FVS,mildenhall2021nerf,lindell2022bacon,muller2022instant,barron2021mip, yu2021pixelnerf} have drawn from work on scene representations and neural rendering~\cite{tewari2022advances}. 
While they achieve impressive results on interpolating near input views, most methods are trained purely with regression objectives and struggle in dealing with ambiguity or longer-range extrapolations. 
When challenged with the task of novel view synthesis from sparse inputs, they can only tackle mildly ambiguous cases, \textit{i.e.}, cases where the conditional distribution of novel renderings is well approximated by the mean estimator of this distribution\,---\,obtained by minimizing a pixel-wise L1 or L2 loss~\cite{zhang2016colorful, sitzmann2019scene, yu2021pixelnerf}. 
However, in highly ambiguous cases, for example when parts of the scene are occluded in all the given views, the conditional distribution of novel renderings becomes multi-modal and the mean estimator produces blurry novel views (see Fig.~\ref{fig:generative_motivation}). Because of these limitations, regression-based approaches are limited to short-range view interpolation of object-centric scenes and struggle in long range extrapolation of unconstrained scenes.

In contrast, generative approaches rely on generative priors and solve the novel view synthesis problem by generating random plausible samples from this conditional distribution. 
Existing generative models for view synthesis~\cite{rombach2021geometryfree, wiles2020synsin, ren2022look, liu2021infinite} autoregressively extrapolate one or a few input images with few or no geometry priors. 
For this reason, most of these methods struggle with generating geometrically consistent sequences\,---\,renderings are only approximately consistent between frames and lack a coherent rigid scene structure. In this work, we present an NVS method that bridges the gap between geometry-based and generative view synthesis approaches for both geometrically consistent and generative rendering.

Our method leverages recent developments in diffusion models. 
Specifically, conditional diffusion models~\cite{saharia2022image, saharia2022palette, ramesh2022hierarchical, rombach2022high, saharia2022photorealistic} can be directly applied to the task of NVS. Conditioned on input images, these models can sample from the conditional distribution of output renderings. As a generative model, they naturally handle ambiguity and lend themselves to continued autoregressive extrapolation of plausible outputs. However, as we show in Sec.~\ref{sec:Experiments} (Tab.~\ref{Tab:ShapeNet}), an image diffusion framework alone struggles to synthesize 3D-consistent views. 

Geometry priors remain valuable for ensuring view consistency when operating on complex scenes, and pixel-aligned features~\cite{saito2019pifu, yu2021pixelnerf, wang2021ibrnet} have been shown to be successful for conditioning scene representations on images. We incorporate these ideas into the architecture of our diffusion-based NVS model with the inclusion of a latent 3D feature field and neural feature rendering~\cite{niemeyer2021giraffe}. Unlike previous view synthesis works that include neural fields, however, our latent feature field captures a distribution of scene representations rather than the representation of a specific scene. A rendering from this latent field is distilled into the rendering of a particular scene realization through diffusion sampling at inference. This novel formulation is able to both handle ambiguity resulting from long-range extrapolation and generate geometrically consistent sequences.


In summary, contributions of our work include: 
\begin{itemize}
    \item We present a novel view synthesis method that extends 2D diffusion models to be 3D-aware by conditioning them on 3D neural features extracted from input image(s).
    \item We demonstrate that our 3D feature-conditioned diffusion model can generate realistic novel views given as little as a single input image on a wide variety of datasets, including object level, room level, and complex real-world.
    \item We show that with our proposed method and sampling strategy, our method can generate long trajectories of realistic, multi-view consistent novel views without suffering from the blurring of regression models or the drift of pure generative models.
\end{itemize}
We will make the code and pre-trained models available.

\section{Related work}
Focusing on novel view synthesis (NVS) from as little as a single image, our work touches on several areas at the intersection of 3D reconstruction, NVS, and generative models.
\vspace{-0.15in}
\paragraph{Geometry-based novel view synthesis.} 
A large body of prior works for NVS recovers the 3D structure of a scene by estimating the input images' camera parameters \cite{snavely2006photo, schonberger2016structure} and running multi-view stereo (MVS) \cite{agarwal2011building, goesele2007multi}. The recovered explicit geometry proxies enable NVS
but fail to synthesize photorealistic and complete novel views especially for occluded regions. Some recent methods \cite{Riegler2020FVS,Riegler2021SVS} combine 3D geometry from an MVS pipeline with deep learning--based NVS, but the overall quality may suffer if the MVS pipeline fails.  Other explicit geometric representations, such as depth maps~\cite{flynn2016deepstereo,single_view_mpi}, multi-plane images~\cite{flynn2019deepview,zhou2018stereo}, or voxels~\cite{sitzmann2019deepvoxels,Lombardi:2019} are also used by many recent NVS approaches, as surveyed by Tewari et al.~\cite{tewari2022advances}.

\vspace{-0.15in}

\paragraph{Regression-based novel view synthesis.}
Many deep learning--based approaches to NVS are supervised to predict training views with regression. These works often employ 3D representations for scenes and differentiable neural rendering~\cite{sitzmann2019scene, mildenhall2021nerf}. While many methods are optimized on a per-scene basis with dense input views~\cite{mildenhall2021nerf}, few-shot NVS approaches are designed to generalize across a class of 3D scenes, which enable them to make predictions from one or a few input images at inference. Among few-shot NVS methods, some rely on test-time optimization~\cite{sitzmann2019scene, jang2021codenerf} or meta learning~\cite{sitzmann2020metasdf, tancik2021learned}, while others lift input observations via encoders~\cite{single_view_mpi, niklaus20193d, zhou2018stereo, yu2021pixelnerf, grf2020, mvsnerf, wang2021ibrnet} and predict novel views in a feed-forward fashion. A recent trend has some NVS methods forgoing geometry priors for light fields~\cite{sitzmann2021lfns} or transformers~\cite{sajjadi2022scene, kulhanek2022viewformer}, but these geometry-free methods are otherwise trained similarly to other regression-based NVS algorithms.

\vspace{-0.15in}

\paragraph{Generative models for novel view synthesis.}
A separate line of work studies methods for long-range view extrapolation. Because venturing far beyond the observed views requires generating parts of the scene, these methods are typically grounded in generative models. A common thread amongst these methods is that they often contain only weak geometry priors, e.g., sparse feature point clouds~\cite{wiles2020synsin, rockwell2021pixelsynth, koh2021pathdreamer}, or lack geometry priors altogether~\cite{rombach2021geometryfree, ren2022look}. As image-translation-based generative models, they are capable of conditioning on their own previous generations to autoregressively synthesize long camera trajectories, sometimes infinitely~\cite{liu2021infinite, li2022infinitenature}. Because the focus is on extrapolating at large scales, these methods ordinarily achieve only approximate view consistency at longer ranges.

\vspace{-0.15in}

\paragraph{3D GANs.} 3D GANs \cite{nguyen2019hologan,schwarz2020graf,chan2021pi,chan2022efficient,gu2021stylenerf,or2022stylesdf, xu20223d, zhou2021cips,epigraf,xue2022giraffehd,zhang2022mvcgan,deng2022gram,xiang2022gramhd,bergman2022gnarf} combine an adversarial~\cite{goodfellow2020generative} training strategy with implicit neural scene representations to learn generative models for 3D objects. While typically tasked with unconditional synthesis of 3D objects, a trained 3D GAN contains a strong prior for 3D shapes and can be inverted for NVS of detailed scenes~\cite{chan2021pi, chan2022efficient}. 
3D GANs have been extensively developed to achieve compositionality~\cite{niemeyer2021giraffe}, higher rendering resolution~\cite{chan2022efficient, gu2021stylenerf, epigraf}, video generation~\cite{bahmani20223d}, and scalability to larger scenes~\cite{devries2021unconstrained}. GANs, however, are notoriously difficult to train, and their 3D inversions from an input image are often brittle without additional 3D priors~\cite{xie2022high} or an accurate camera input~\cite{ko20233d3dganinversion}.
Moreover, most 3D GANs assume canonical camera poses and limit their optimal operating ranges to single objects. 

\vspace{-0.15in}

\paragraph{2D diffusion models.}
2D diffusion models~\cite{ho2020denoising, sohl2015deep, song2019generative, karras2022elucidating} have transformed image synthesis. Favorable properties such as mode coverage and a stable training objective have enabled them to outperform~\cite{dhariwal2021diffusion} previous generative models~\cite{goodfellow2020generative} on unconditional generation. Diffusion models have also been shown to be excellent at modeling conditional distributions of images, where the conditioning information may be a class label~\cite{song2020score, dhariwal2021diffusion}, text~\cite{ramesh2022hierarchical, rombach2022high, saharia2022photorealistic} or another image~\cite{ho2022cascaded, saharia2022image, saharia2022palette, chandeep}.

\vspace{-0.15in}

\paragraph{Recent 3D diffusion works.}
Recently, DreamFusion \cite{poole2022dreamfusion} and 3DiM \cite{watson2022novel} apply 2D image diffusion models to build 3D generative models. DreamFusion performs text-guided 3D generation by optimizing a NeRF from scratch. 3DiM performs novel view synthesis conditioned on input images and poses (similar to \cite{ren2022look}) and does not employ any explicit geometry priors; it aggregates multiple observations at inference using a unique stochastic conditioning scheme. By contrast, the geometry priors present in our approach enable 3D consistency with a much lighter-weight model (90M for ours vs 471M or 1.3B for 3DiM~\cite{watson2022novel}), and because our model naturally handles multiple input views, we have the flexibility to choose efficient sampling schemes at inference. While code for 3DiM is unavailable, we compare to a similar geometry-free variant in Sec.~\ref{sec:Experiments} (Tab.~\ref{Tab:ShapeNet}) and to stochastic view conditioning in the supplement.

\section{Method}
\label{sec:Method}

\newcommand{\cam}{\mathbf{P}}
\newcommand{\im}{\boldsymbol{x}}
\newcommand{\camgt}{\mathbf{P}^{\text{input}}}
\newcommand{\camgts}{\mathbf{P}^{\text{inputs}}}
\newcommand{\imgt}{\boldsymbol{x}^{\text{input}}}
\newcommand{\imgts}{\boldsymbol{x}^{\text{inputs}}}
\newcommand{\camquery}{\mathbf{P}^{\text{target}}}
\newcommand{\impred}{\boldsymbol{x}^{\text{target}}}
\newcommand{\noisytarget}{\boldsymbol{y}^{\text{target}}}
\newcommand{\rendering}{\Gamma}
\newcommand{\noisyimg}{\boldsymbol{y}}

\newcommand{\xinput}{\boldsymbol{x}^{\text{input}}}
\newcommand{\xtarget}{\boldsymbol{x}^{\text{target}}}
\newcommand{\caminput}{\mathbf{P}^{\text{input}}}
\newcommand{\camtarget}{\mathbf{P}^{\text{target}}}
\newcommand{\denoiser}{D}
\newcommand{\translation}{T}
\newcommand{\unet}{U}
\newcommand{\feature}{F}

Here we describe the architecture of our NVS model for both single and multiple-view conditioning, and we explain our training and inference methods.

In novel view synthesis, we are given a set of input images $\imgts$ and camera parameters $\camgts$ with associated pose and intrinsics and are tasked with making a prediction for a query view given a set of query camera parameters.

Our goal is to sample novel views from the corresponding conditional distribution:
\begin{equation} \label{eq:conditional_distribution}
    p(\impred | \imgts, \camgts, \camquery) \text{.}
\end{equation}

\begin{figure}
    \centering
    \includegraphics[width=\columnwidth]{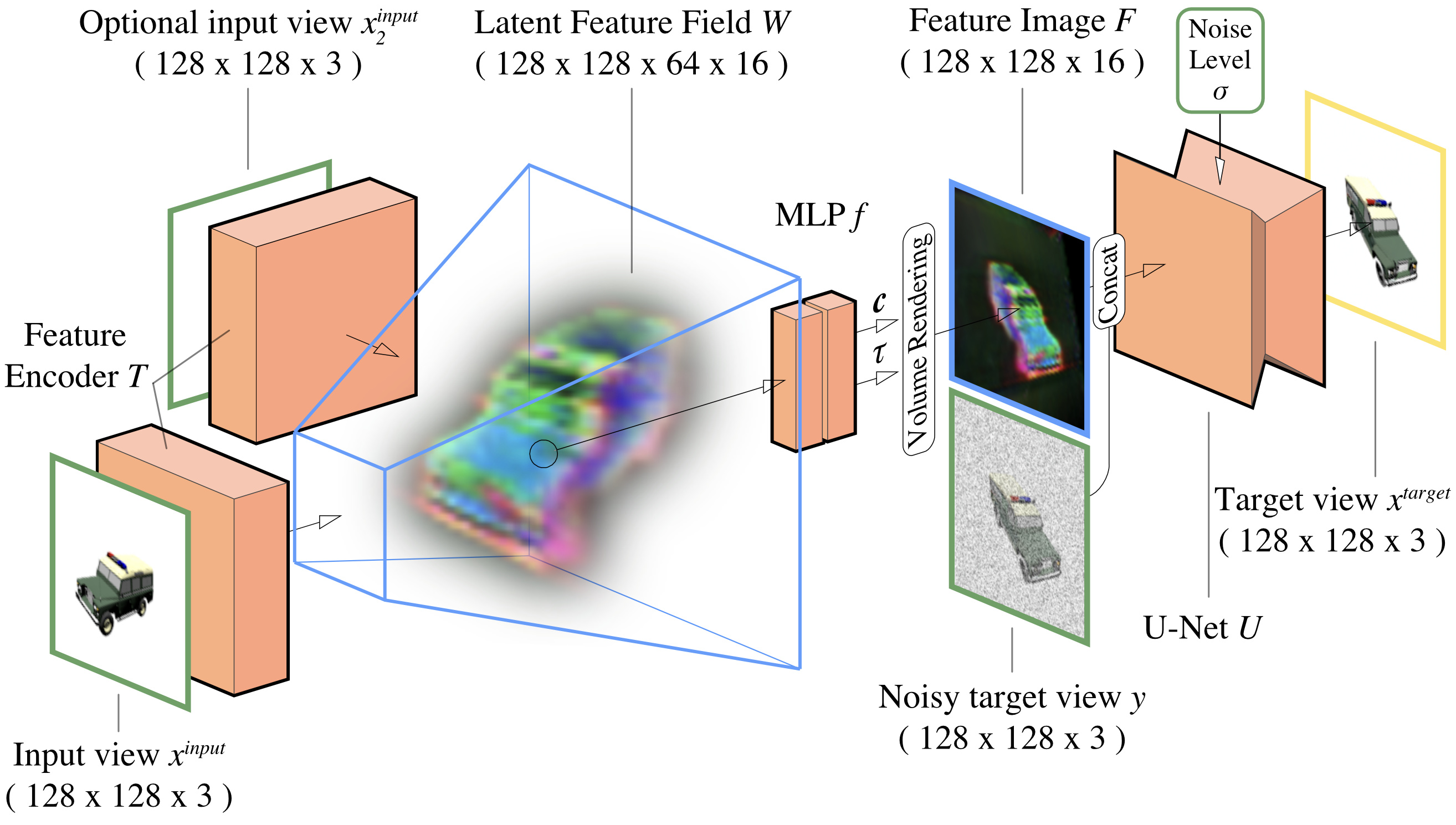}
    \caption{
    Illustration of our framework $\denoiser$.
    The pipeline receives as input one or more input views $\boldsymbol{x}$ and the camera parameters associated with input and target views.
    We extract features from each input view $\boldsymbol{x}$ using $T$ and unproject them into a feature volume $\boldsymbol{W}$.
    These volumes are aggregated using a mean-pooling operation, decoded by a small MLP $f$, and a feature image $\feature$ is created by projecting into the target view $\impred$ using volume rendering.
    The U-Net denoiser $\unet$ then takes in the resulting feature image $\feature$ as well as a noisy image of the target view $\boldsymbol{y}$ and noise level $\sigma$, and produces a denoised image of the target view $\impred$.
    }
    \label{fig:network}
\end{figure}

\subsection{3D-aware diffusion model architecture}

Diffusion models rely on a denoiser trained to predict $\mathbb{E}_{p(\boldsymbol{x}|\boldsymbol{y})}[\boldsymbol{x}]$ given $\boldsymbol{y}$, a noisy version of $\boldsymbol{x}$ with noise standard deviation $\sigma$. An image is generated by drawing $\boldsymbol{y}_0 \sim \mathcal{N}(0, \sigma_\text{max}^2 \mathbf{I})$ and iteratively denoising it according to a sequence of noise levels $\sigma_0 = \sigma_\text{max} > ... > \sigma_N = 0$. 

 In our work, we directly repurpose 2D diffusion models to model the distribution in Eq.~\ref{eq:conditional_distribution}. The intuition is that generative novel view synthesis is identical to any other conditional image generation task\,---\,all we need to do is condition a 2D image diffusion model on the input image and the relative camera pose. However, while there are many ways of applying this conditioning, some may be more effective than others (see Tab.~\ref{Tab:ShapeNet} and ablation studies of different options in Sec.~\ref{sec:ablation}). By incorporating geometry priors 
 in the form of a 3D feature field and neural rendering, we give our architecture a strong inductive bias towards geometrical consistency.
 
 Fig.~\ref{fig:network} summarizes the design of our conditional-desnoiser-based pipeline $\denoiser$ that takes as inputs a noisy target view $\noisyimg$, conditioning information $(\imgts, \camgts, \camtarget)$ and a noise level $\sigma$.
Our strategy builds upon pixel-aligned implicit functions~\cite{saito2019pifu, yu2021pixelnerf} and neural rendering. Following Fig.~\ref{fig:network}, given a single input image $\boldsymbol{x}$ taken from an input view camera $\mathbf{P}$, we use an image-to-image translation network $\translation$ to predict a feature image with $c \times d$ channels and reshape it into a feature volume $\boldsymbol{W}$ that spans the source camera frustum. 
$d$ then corresponds to the depth dimension of the volume and $c$ to the number of channels in each cell of the volume (typically, $c=16$ and $d=64$). Given a query camera $\camtarget$, we cast rays in 3D space. Continuing on Fig.~\ref{fig:network}, for any point $\mathbf{r}$ along a ray, we sample the volume $\boldsymbol{W}$ with trilinear interpolation and decode the obtained feature $w=\boldsymbol{W}(\mathbf{r})$ with a small multi-layer perceptron (MLP) $f$ to obtain a density $\tau$ and a feature vector $\mathbf{c}$
\begin{equation}
    \left( \tau, \mathbf{c} \right) = f(w).
\label{eq:rendering}
\end{equation}
By projecting this feature field into the target view using volume rendering~\cite{max1995optical,mildenhall2021nerf}, we obtain a feature image $F$ in Fig.~\ref{fig:network}:
\begin{equation}
    \feature(\boldsymbol{x}, \mathbf{P}, \camtarget) = \text{{\sc render}}(f\circ\translation(\boldsymbol{x}), \mathbf{P}, \camtarget).
\label{eq:encoder}
\end{equation}

In practice, we employ the image segmentation architecture \textit{DeepLabV3+}~\cite{chen2018encoder, Iakubovskii:2019} for $\translation$, and implement $f$ as a two-layer ReLU MLP with 64 channels. We perform volume rendering over features in the same way as \emph{NeRF}~\cite{mildenhall2021nerf}. We use input/output image resolution $128^2$ in all experiments.

The feature image $\feature$ is concatenated to the noisy image $\boldsymbol{y}$ and passed as input to a denoiser network $\unet$ to produce the final target view $\impred$ (see Fig.~\ref{fig:network}). 
We use \emph{DDPM++}~\cite{song2020score, karras2022elucidating} for $\unet$, where
%
%
\begin{equation} \label{eq:pipeline}
    \denoiser(\noisyimg~; \imgts, \camgts, \camtarget, \sigma) = \unet(\noisyimg, \feature; \sigma)
\end{equation}

Fig.~\ref{fig:network} and Eq.~\ref{eq:pipeline} summarize the design of $\denoiser$. The total number of trainable parameters in $\denoiser$ is 90M.

\subsection{Incorporating multiple views}
\label{sec:incorporating_multiple_views}

The previous section describes our approach to conditioning on a single input view. However, additional information in the form of multiple input views reduces uncertainty and enables our model to sample renderings from a narrower distribution. When multiple conditioning views are available, we process each input image independently into a separate feature volume.

Eq.~\ref{eq:rendering} can be generalized to $n$ conditioning views by averaging the features $w_j=\boldsymbol{W}_j(\mathbf{r})$ obtained for each input image $\boldsymbol{x}_j$, as in~\cite{yu2021pixelnerf}:
\begin{equation}
    \left(\tau, \mathbf{c}\right) = f\left(\frac{1}{n}\sum_{j=1}^n w_j\right).
\label{eq:multi-rendering}
\end{equation}

To leverage this strategy during inference, we train our model by conditioning with multiple (variable) input images. Conditioning using multiple input images helps to ensure smooth, loop-consistent video synthesis. While conditioning on only the previous frame is sufficient for view consistency in a small view change, it does not guarantee loop closure. In practice, we find that conditioning on a subset of previous views helps to enforce correct loop closure while maintaining reasonable view to view consistency.

\subsection{Training}

At each iteration during training, we sample a batch of target images, input images, and their associated camera poses, where the targets and inputs are constrained to be from the same scene. Our model is trained end-to-end from scratch to minimize the following objective
\newcommand{\ptarget}{\mathbf{P}^{\text{target}}}
\newcommand{\pinput}{\mathbf{P}^{\text{input}}}
\begin{align}
    & L \coloneqq
    \mathbb{E}_{(\xtarget, \imgts, \camquery, \camgts) \sim p_\text{data}}
    \mathbb{E}_{\varepsilon \sim \mathcal{N}(0, \sigma^2 \mathbf{I})} \label{eq:objective} \\
    & \left[ \|\denoiser(\xtarget + \varepsilon~; \imgts,\camgts,\camquery, \sigma) - \xtarget \|_2^2 \right], \nonumber
\end{align}
where $\sigma$ is sampled during training according to the strategy proposed by \textit{EDM} \cite{karras2022elucidating}. The number of conditioning views for a query is drawn uniformly from $\{1, 2, 3\}$ at every iteration. During training, we apply non-leaking augmentation~\cite{karras2022elucidating} to $\unet$ and augment input images with small amounts of random noise. Please see the supplement for hyperparameters and additional training details.

\subsection{Generating novel views at inference}
\label{sec:methods_inference}
Sampling a novel view with our method is identical to sampling an image with a conditional diffusion model. The specific update rule for the denoised image is determined by the choice of sampler. In our experiments, we use a deterministic 2\textsuperscript{nd} order sampling strategy \cite{karras2022elucidating}, with 25 or fewer denoising steps. Other sampling strategies~\cite{song2020score, song2020denoising} can be dropped in if other properties (\textit{e.g.}, stochastic sampling) are desired.

In order to improve efficiency at inference, we decouple $\rendering$ and $\unet$. Rather than running both $\rendering$ and $\unet$ at every step during sampling, we first render the feature image $\feature$ as a preprocessing step and reuse it for each iteration of the sampling loop -- while $\unet$ must run every step during inference, $\rendering$ is run only once.
\vspace{-0.15in}

\paragraph{Alternative ``one-step'' inference.}
An alternative variant of our model to generating an image with iterative denoising is to produce the image with a single step of denoising. Intuitively, the one-step prediction of a model trained with Eq.~\ref{eq:objective} should behave identically to the prediction of a model trained to minimize pixel-wise MSE. Thus, this alternative inference mode is representative of regression-based methods. A model trained as described is capable of both generative sampling and deterministic one-step inference—no architecture or training modifications are required.



\subsection{Autoregressive generation}
In order to generate consistent sequences, we take an autoregressive approach to synthesizing sequential frames. Instead of independently generating each frame conditioned only on the input images, which would lead to large deviations between frames, we generate each frame conditioned on the inputs as well as a subset of previously generated frames. While there are many possible ways of selecting conditioning views, a reasonable setting that we use in our experiments is to condition on the input image(s), the most recently generated image, and five additional images drawn at random from the set of previously generated frames.

We found this default conditioning setting to be a good starting point that balances short range, frame-to-frame consistency, long-range consistency across the scene, and compute cost, but other variants may be preferred to emphasize specific qualities.

While one might expect errors and artifacts to accumulate throughout long autoregressive sequences, in practice we find that our model effectively suppresses such errors, making it suitable for extended sequence generation. Please see the supplement for alternative autoregressive schemes.

\section{Experiments}
\label{sec:Experiments}

\begin{figure}[t!]
    \centering
    \includegraphics[width=\linewidth]{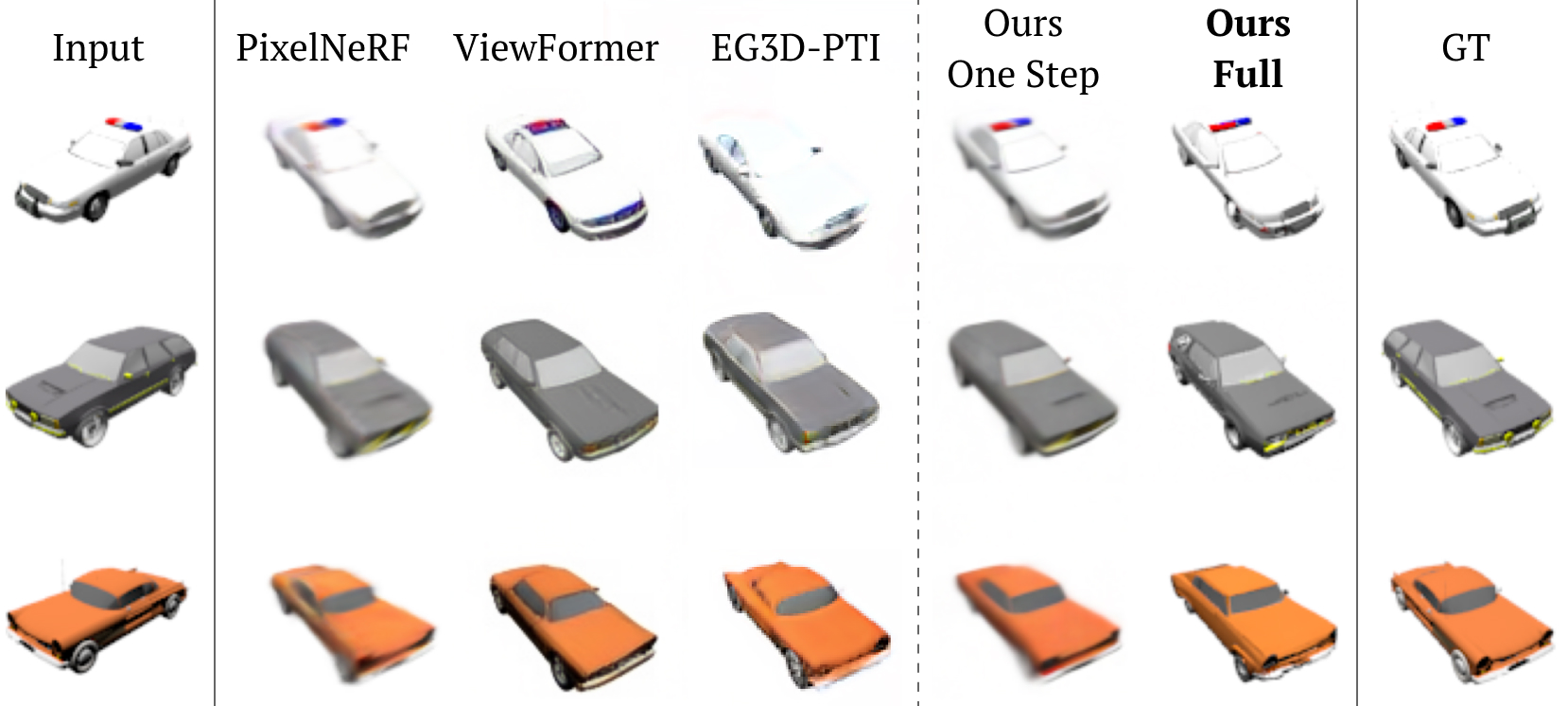}
    \caption{
      Qualitative comparison on ShapeNet~\cite{chang2015shapenet} with one input view. Unlike regression-based approaches, our method produces sharp realizations. With one-step inference, our approach behaves like a mean estimator of the novel view, similarly to PixelNeRF.
    }
    \label{fig:shapenet_qualitative}
    \vspace{-5pt}
\end{figure}

We evaluate the performance of our generative NVS method on ShapeNet~\cite{chang2015shapenet} ``cars" and Matterport3D~\cite{chang2017matterport3d}, two starkly different datasets. ShapeNet is representative of synthetic, object-centric datasets that have long been dominated by regression-based approaches to NVS (e.g.,~\cite{yu2021pixelnerf, sitzmann2021lfns}). Meanwhile, long-range NVS on Matterport3D is prototypical of unbounded scene exploration, where generative models with weak geometry priors~\cite{wiles2020synsin, rombach2021geometryfree, ren2022look} have seen more success. Finally, we stress-test our method on the challenging Common Objects in 3D (CO3D) \cite{reizenstein2021common}, an unconstrained real-world dataset\,---\,to our knowledge, our work is the first to attempt single-shot NVS on this dataset while including its complex backgrounds.
Our method improves upon the state-of-the-art for all tasks. For additional results, please refer to the videos contained in the supplement.
\vspace{-0.15in}

\paragraph{Baselines and implementation details.} For ShapeNet and CO3D, we compare our method to PixelNeRF~\cite{yu2021pixelnerf}, a state-of-the-art NeRF-based method for NVS, and ViewFormer~\cite{kulhanek2022viewformer}, a transformer-based, geometry-free approach to NVS. For ShapeNet, we additionally provide a comparison with EG3D-PTI~\cite{chan2022efficient}, which is based on a state-of-the-art 3D GAN for object-scale scenes, and a numerical comparison with 3DiM~\cite{watson2022novel}, a recent geometry-free diffusion method for NVS. For Matterport3D, we compare our method against the state-of-the-art on this dataset: Look Outside The Room~\cite{ren2022look}, a transformer-based, geometry-free NVS method designed for room-scale scenes, and to additional SOTA methods, including SynSin~\cite{wiles2020synsin} and GeoGPT~\cite{rombach2021geometryfree} in Tab.~\ref{Tab:Matterport}. 

\begin{table}[t!]
\centering
\resizebox{\columnwidth}{!}{%
\begin{tabular}{@{}lllllll@{}}
\hline
& & FID$\downarrow$ $\!\!\!\!$& $\!\!$LPIPS$\downarrow$ $\!\!\!\!$&$\!\!\!$ DISTS$\downarrow$ $\!\!\!\!$ & $\!\!\!\!$ PSNR$\uparrow$ $\!\!\!\!\!\!$ & SSIM $\uparrow$ \\
\hline
\multicolumn{2}{l}{$\!\!\!\!$PixelNeRF~\cite{yu2021pixelnerf}}
& 65.83 & 0.146 & 0.203 & \textbf{23.2} & 0.90\\
\multicolumn{2}{l}{$\!\!\!\!$ViewFormer~\cite{kulhanek2022viewformer}$\!\!\!\!$}
& 20.82 & 0.146 & 0.161 & 19.0 & 0.83 \\
\multicolumn{2}{l}{$\!\!\!\!$EG3D-PTI~\cite{chan2022efficient}$\!\!\!\!$}
& 27.23 & 0.150 & 0.310 & 19.0 & 0.85 \\
\multicolumn{2}{l}{$\!\!\!\!$3DiM (autoregressive)~\cite{watson2022novel}\textsuperscript{\dag}$\!\!\!\!$}
& 8.99 &  &  & 21.01 & 0.57 \\
\hline
 \parbox[t]{0mm}{\multirow{4}{*}{\rotatebox[origin=c]{90}{Ours}}} & $\!\!$ Explicit & 8.09 & 0.129 & 0.158 & 19.1 & 0.86 \\
& $\!\!$ Geom.-Free\!\! & 16.68 & 0.342 & 0.329 & 13.1 & 0.74 \\
& $\!\!$ One-Step      & 42.07 & 0.150 & 0.178 & \textbf{23.2} & \textbf{0.91}\\
& $\!\!$ Full (autoregressive) & 11.08 & 0.120 & 0.146 & 20.6 & 0.89 \\ 
& $\!\!$ Full & \textbf{6.47} & \textbf{0.104} & \textbf{0.145} & 20.7 & 0.89 \\ 
\hline
\end{tabular}
}
\vspace{3pt}
\caption{  Quantitative comparison of single-view novel view synthesis on ShapeNet cars~\cite{chang2015shapenet, sitzmann2019scene}. \textsuperscript{\dag} As reported by~\cite{watson2022novel}.
}\label{Tab:ShapeNet}
\vspace{-5pt}
\end{table}


\vspace{-0.15in}

\paragraph{Metrics.}
We evaluate the task of novel view synthesis along three axes: ability to (1) recreate the image quality and diversity of the ground truth dataset, (2) generate novel views consistent with the ground truth, and (3) generate sequences that are geometrically consistent. For (1), we use distribution-comparison metrics, FID~\cite{heusel2017fid} and KID~\cite{binkowski2018demystifying}, which are commonly used to evaluate generative models for image synthesis. For (2), we use perceptual metrics LPIPS~\cite{zhang2018perceptual} and DISTS~\cite{ding2020iqa}, which measure structural and texture similarity between the synthesized novel view and ground-truth novel view.
For completeness, we include PSNR and SSIM, although the drawbacks of these metrics are well-studied: these raw pixel metrics have been shown to be poor evaluators of generative models as they favor conservative, blurry estimates that lack detail~\cite{saharia2022image, saharia2022palette}.
For (3), we provide COLMAP~\cite{schoenberger2016sfm, schoenberger2016mvs} reconstructions of generated video sequences, a standard evaluation for 3D consistency in 3D GANs~\cite{schwarz2020graf, chan2021pi, chan2022efficient}. Dense, well-defined point clouds are indicative of geometrically consistent frames. We calculate Chamfer distances between reconstructions of the ground-truth images and reconstructions of generated sequences to quantitatively evaluate geometrical consistency.

\begin{figure}
    \centering
    \includegraphics[width=\linewidth]{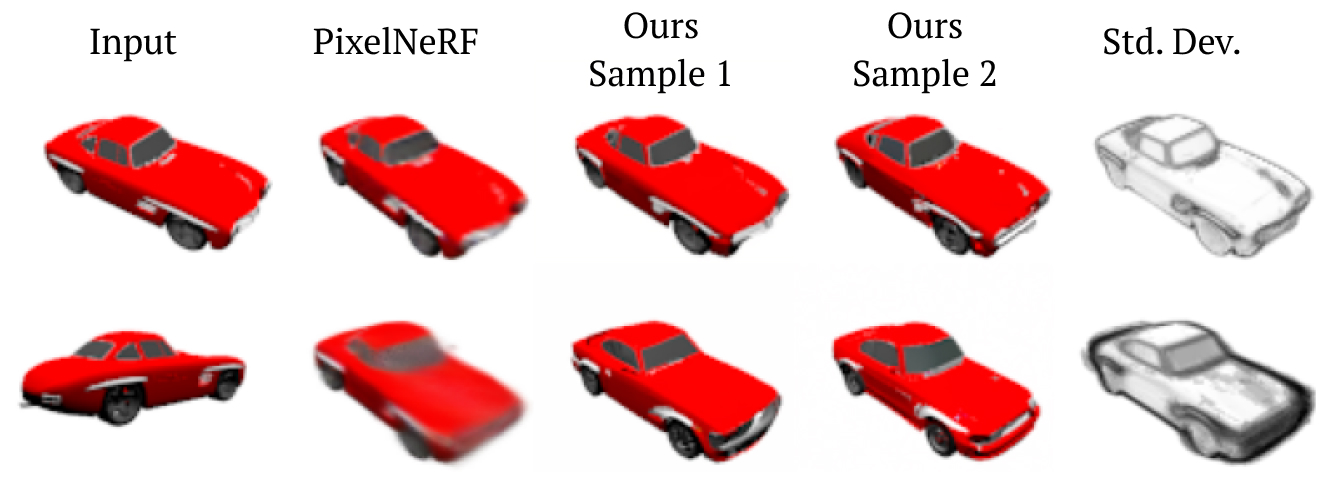}
    \caption{Generating new views from more (bottom) or less (top) ambiguous conditioning information. PixelNeRF~\cite{yu2021pixelnerf} is constrained to output deterministic novel views and renders an average of all plausible renderings that are consistent with the input view. In comparison, our method samples the conditional distribution, leading to sharp but different realizations. In the last column, we show the per-pixel standard deviation of the novel view and show that unseen areas are more ambiguous, \textit{i.e.}, vary more from one sample to the other. Pixel-wise standard deviation is computed over $50$ samples. Dark pixels indicate higher ambiguity.}
    \label{fig:ambiguity}
    \vspace{-5pt}
\end{figure}

\begin{figure}
    \centering
    \includegraphics[width=\linewidth]{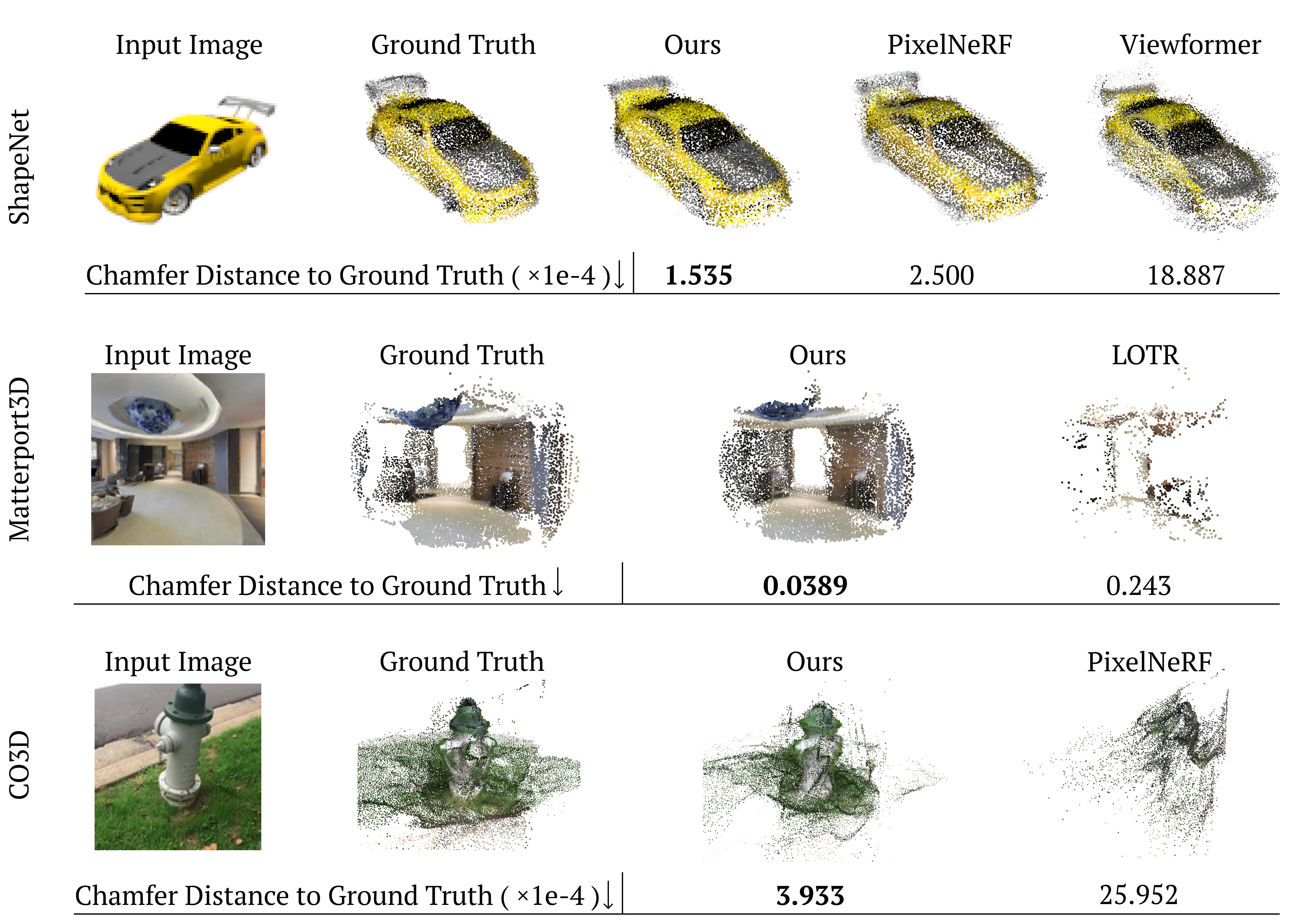}
    \caption{
    COLMAP reconstructions from video sequences produced by our method are dense, well-defined, and highly similar to reconstructions of the ground-truth images, demonstrating a high degree of geometric consistency, as measured by Chamfer distance. The three rows show results on ShapeNet, Matterport3D, and CO3D, respectively.
    }
    \label{fig:all_colmap}
    \vspace{-5pt}
\end{figure}

\subsection{ShapeNet}
\label{sec:ShapeNet}
We standardize our training and evaluation on the single-class, single-view NVS benchmark described in~\cite{yu2021pixelnerf,sitzmann2019scene,kulhanek2022viewformer}. The ShapeNet training set contains 2,458 cars, each with 50 renderings randomly distributed on the surface of a sphere. For evaluation, we use the provided test set with 704 cars, each with 250 rendered images and poses on an Archimedean spiral. All evaluations are conducted with a single input image. For our model, we evaluate both independently generated frames and frames generated with autoregressive conditioning. In addition to our model and the baselines, we provide additional comparisons to several ablative variants of our approach, which are discussed in more detail in Sec.~\ref{sec:ablation}.

Fig.~\ref{fig:shapenet_qualitative} provides a qualitative comparison against baselines for single-view novel view synthesis on ShapeNet. In contrast to PixelNeRF, which predicts a blurry mean of the conditional distribution, our method (Ours Full) generates sharp realizations. While ViewFormer also produces sharp images due to training with a perceptual loss, its renderings fail to transfer some small details, such as headlight shape, from the input.

In Tab.~\ref{Tab:ShapeNet}, we report the quality of novel renderings produced by our method and baselines, as measured by FID~\cite{heusel2017fid}, LPIPS~\cite{zhang2018perceptual}, DISTS~\cite{ding2020image}, PSNR, and SSIM~\cite{wang2004image}. As a generative model, our method creates sharp, diverse outputs, which closely match the image distribution; it thus scores more favorably in FID than regression baselines~\cite{yu2021pixelnerf, kulhanek2022viewformer}, which tend to produce less finely detailed renderings. Our method outperforms baselines in LPIPS and DISTS, which indicates that our method produces novel views that achieve greater structural and textural similarity to the ground truth novel views. We would not expect a generative model to outperform a regression model in PSNR and SSIM, and indeed, renderings from PixelNeRF achieve higher scores in these pixel-wise metrics than realizations from our model. However, we note that the one-step denoised prediction of our model (described in Sec.~\ref{sec:methods_inference}) is able to match PixelNeRF's state-of-the-art PSNR and SSIM. While our method with autoregressive conditioning does not surpass 3DiM~\cite{watson2022novel}, it achieves competetive scores with a lighter weight model (90M vs 471M params) and fewer diffusion steps (25 vs 512).

In Fig.~\ref{fig:ambiguity}, we demonstrate that for a given observation, our model is capable of producing multiple plausible realizations. When conditioning information is reliable, such as when the query view is close to the input view, ambiguity is low and samples are drawn from a narrow conditional distribution. For more ambiguous inputs, such as when the model is tasked with recreating regions that were occluded in the input image, our model produces plausible realizations with more variation. In contrast, regression-based methods such as PixelNeRF deterministically predict the mean of the conditional distribution and are therefore unable to create high quality realizations when the target view is far from conditioning information and the conditional distribution is large.

Fig.~\ref{fig:all_colmap} shows that our method can also achieve high geometrical consistency when combined with autoregressive generation as validated by dense point cloud reconstruction and the Chamfer distance to the ground truth.


\subsection{Matterport3D}
\label{sec:MP3D}
Beyond ShapeNet, we seek to show the effectiveness of our method on the Matterport3D (MP3D) dataset that features building-scale, real-world scans. We use the provided code of \cite{ren2022look} to sample trajectories of embodied agents and generate 6,000 videos for training and 200 videos for testing, using the provided 61/18 training and test splits.
We train our model by sampling random pairs of input and target images from the same video sequence, where 50\% of input views are drawn from within ten frames of the target view and the rest are sampled randomly from the video sequence. The rest of the training procedure is equivalent to the one we use with ShapeNet.

\begin{figure}[t!]
    \centering
    \includegraphics[width=\linewidth]{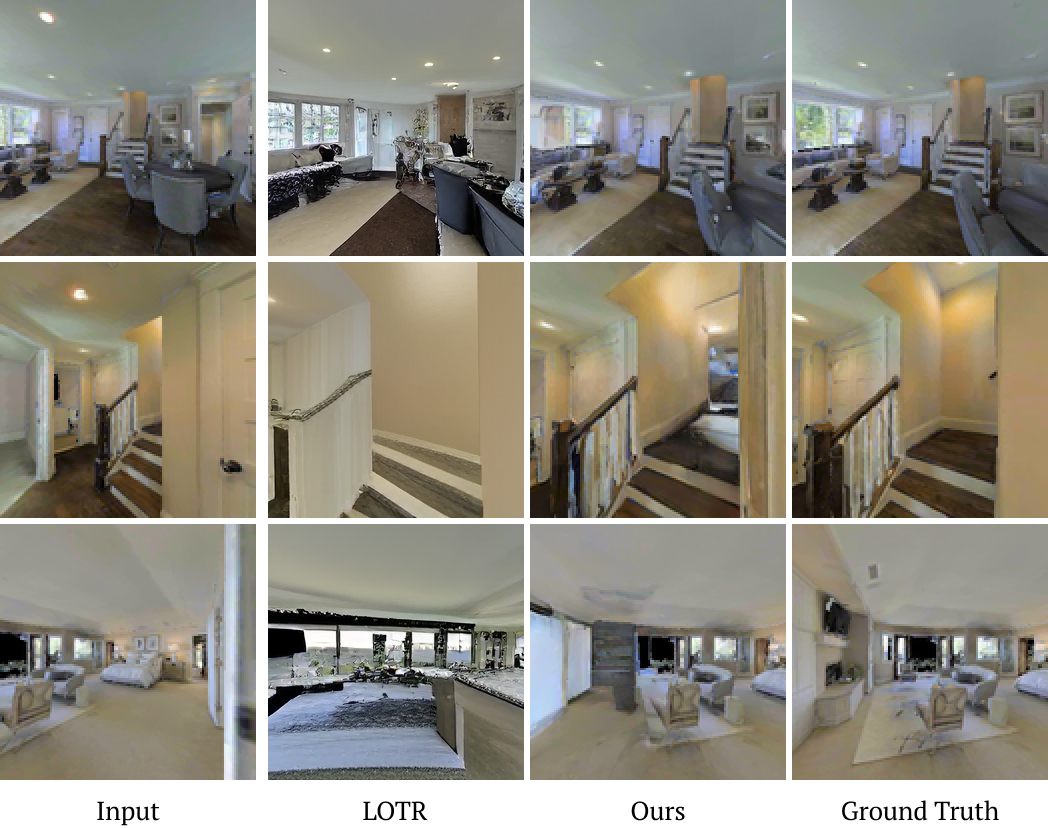}
    \caption{Qualitative comparison on Matterport3D~\cite{chang2017matterport3d} for NVS. Given a single input image (1st col.), we autoregressively run our method and LOTR~\cite{ren2022look} for 10 frames to synthesize novel view images (2\textsuperscript{nd} and 3\textsuperscript{rd} columns). Ground truth images for the corresponding query camera poses are shown in the fourth column. Best viewed zoomed-in.}
    \label{fig:matterport3d_qualitative}
    \vspace{-5pt}
\end{figure}

\begin{table}
\setlength{\tabcolsep}{4.5pt}
\centering
\resizebox{\columnwidth}{!}{%
\begin{tabular}{@{}llllll@{}}
\hline
 & KID$\downarrow$ & $\!\!\!$LPIPS$\downarrow$ $\!\!$& $\!\!\!$DISTS$\downarrow$ $\!\!$& $\!$PSNR$\uparrow$ & $\!\!$SSIM$\uparrow$\\
\hline
LOTR~\cite{ren2022look} (10 f.) & 0.050       &  0.33    &  0.27     & 16.57  &  0.49   \\
Ours (10 f.)                  & \textbf{0.002}      &  \textbf{0.14}    &  \textbf{0.14}     & \textbf{20.80}  &\textbf{0.71}  \\
\hline
SynSin-6X$^*$~\cite{wiles2020synsin}&  0.072   &   0.48    &   0.34    &   14.89   &   0.41 \\
GeoGPT$^*$~\cite{rombach2021geometryfree}& 0.039   &   0.33    &   0.27    &   16.47   &   0.49 \\
LOTR~\cite{ren2022look}& 0.027       &  0.25    &  0.22     & 18.00  &  0.55   \\
Ours& \textbf{0.002}      &  \textbf{0.09}    &  \textbf{0.11}     & \textbf{22.79}  &\textbf{0.79}  \\
\hline
\end{tabular}
}
\vspace{3pt}
\caption{  Quantitative comparison of single-view novel view synthesis on Matterport3D~\cite{chang2017matterport3d}. Here, we use KID since it provides an unbiased estimate when the number of images is small. ``10 f." indicates novel view synthesis for 10 frames from the input image (used 5 frames for the bottom rows). \textsuperscript{*}For SynSin and GeoGPT, we obtained the rendered images from the authors of LOTR. }\label{Tab:Matterport}
\setlength{\tabcolsep}{6pt}
\end{table}


\begin{figure}[t!]
    \centering
    \includegraphics[width=\linewidth]{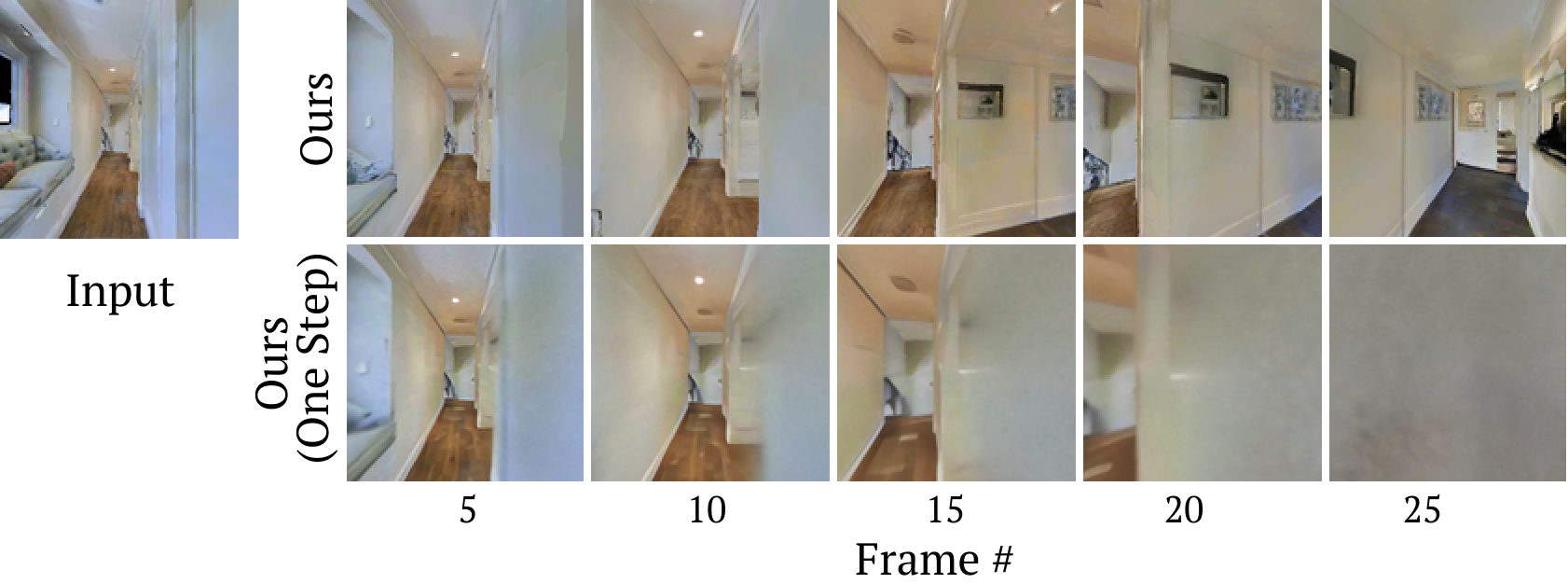}
    \caption{Regression-based models, such as the one-step variant of our approach, struggle to model ambiguity and therefore fail to create plausible renderings far from the input. Generative sampling enables plausible synthesis in ambiguity. When combined with autoregressive generation, we are able to explore areas that were completely occluded in the input.}
    \label{fig:reb_pixelnerf_mp3d}
\end{figure}

\begin{figure}[t!]
    \centering
    \includegraphics[width=\linewidth]{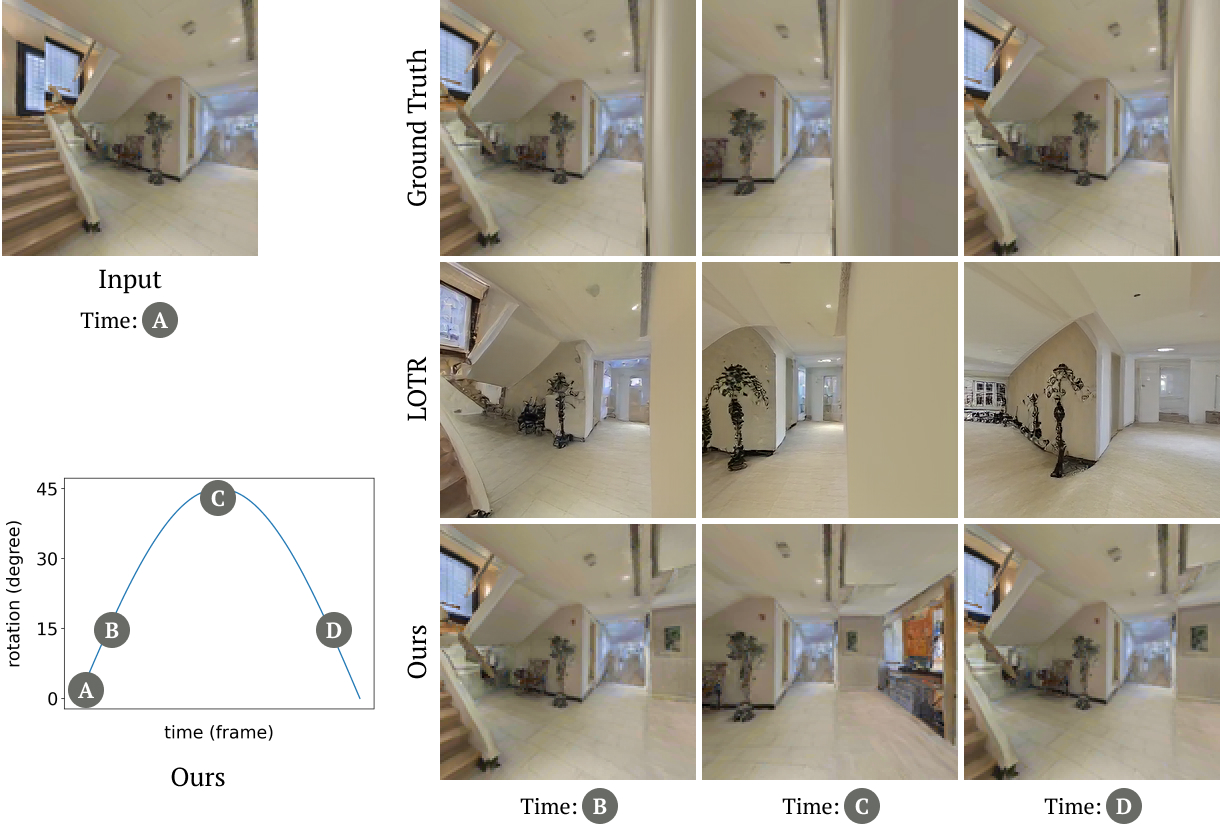}
    \caption{
      Loop closure test on Matterport3D~\cite{chang2017matterport3d}.
      We run our method and LOTR \cite{ren2022look} on a small cyclic rotation angle trajectory (0$^\circ$$\rightarrow$15$^\circ$$\rightarrow$45$^\circ$$\rightarrow$15$^\circ$). 
      Without 3D representations, transformer-based methods, such as LOTR, rely on interpreting raw camera parameters, resulting in weak spatial awareness. Our 3D feature representation more effectively aggregates past observations and provides better loop closure. Best viewed zoomed-in.
    }
    \label{fig:matterport3d_rotate}
    \vspace{-5pt}
\end{figure}

For evaluation, we randomly select an input frame in the test video set (one input frame for each test video), and run ten steps of autoregressive synthesis, following the test camera trajectory; we calculate metrics using all ten synthesized frames. Beyond 10 frames, input and the target frusta rarely overlap, making comparisons against ground truth frames less meaningful. We compare against Look Outside the Room (LOTR)~\cite{ren2022look}, the current state-of-the-art (SOTA) for single-view NVS on Matterport3D that outperforms prior NVS works (i.e., \cite{wiles2020synsin,rockwell2021pixelsynth,rombach2021geometryfree,lai2021video}). We additionally compare against SynSin~\cite{wiles2020synsin} and GeoGPT~\cite{rombach2021geometryfree}, using the 5-frame renderings provided by the authors of LOTR.
Note that, since the trajectories of embodied agents are randomly sampled, the trajectories used for these two baselines are different from those used for our method and LOTR. This comparison measures performance on 200 random trajectories, which is statistically meaningful and the results align with the trends reported in LOTR. 
For all baselines, we downsample the outputs to our output resolution, i.e., $128^2$, and compute the aforementioned metrics against the ground truth images. To measure the realism of the outputs, we choose KID \cite{binkowski2018demystifying}, as it is known to be less biased than FID when the number of test images is small (we use 2000 images). %

The results, summarized in Tab.~\ref{Tab:Matterport}, show that our approach generates novel view predictions that outperform baselines in terms of quality and consistency with the input view. Fig.~\ref{fig:matterport3d_qualitative} supports the trends observed in the metrics—our NVS is noticeably more accurate and realistic than the current SOTA.

In Fig.~\ref{fig:matterport3d_rotate}, we compare against LOTR on a cyclic trajectory. Our method produces better loop closure, indicating higher geometric consistency and showing the effectiveness of incorporating 3D priors. Fig.~\ref{fig:all_colmap} additionally validates the consistency of our results with superior reconstructed point clouds and Chamfer distances.

\subsection{Common Objects in 3D (CO3D)}
We challenge our method with real-world scenes from the Common Objects in 3D (CO3D) \cite{reizenstein2021common} dataset with complete backgrounds. To our knowledge, no prior method has attempted single-shot NVS on CO3D without object masks. We train our method on the hydrant category of the CO3D dataset, which contains 726 RGB videos of real-world fire hydrants. Most videos contain a walkaround trajectory looking in at the hydrant spanning between 60 and 360 degrees, and most videos consist of about 200 frames. We use a 95:5 train/test split to train our model. CO3D is a highly unconstrained and extraordinarily difficult benchmark: scene scale, camera intrinsics, complex backgrounds, and lighting conditions are highly variable between (and sometimes within) scenes. 

\begin{figure}[t]
    \centering
    \includegraphics[width=\linewidth]{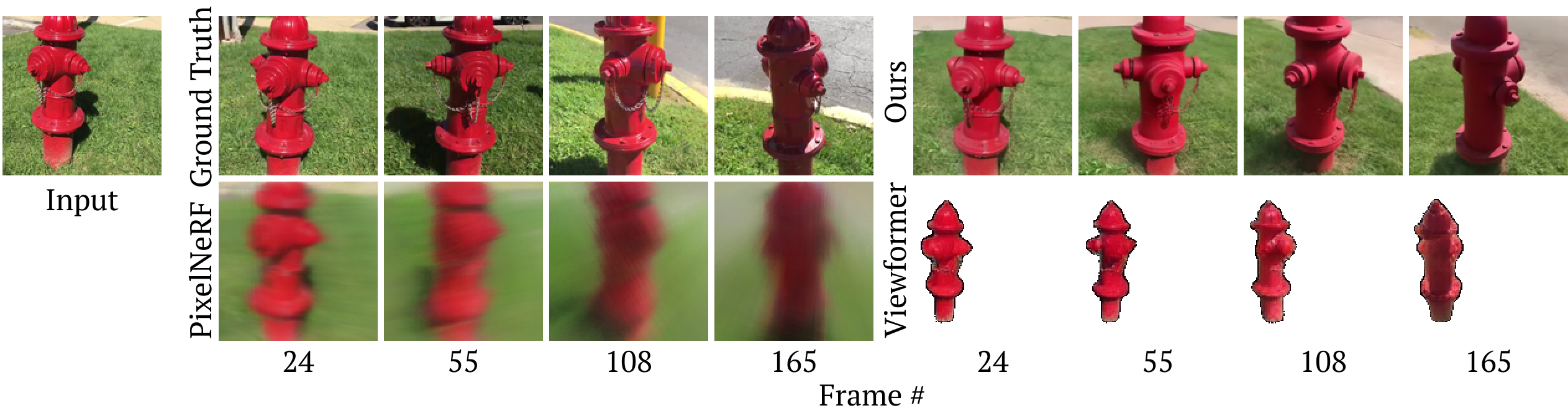}
    \caption{While PixelNeRF produces severe artifacts when the rendering view is far away from the input and ViewFormer requires masks for training on this dataset, our method generates compelling sequences from single-views on challenging, real-world objects of the CO3D dataset~\cite{reizenstein2021common}.}
    \label{fig:CO3D}
\end{figure}

Fig.~\ref{fig:CO3D} compares predictions from our method against baselines on CO3D. Our method produces plausible and sharp foregrounds and backgrounds that do not deteriorate in quality with increasing distance from the source pose.
While we include a qualitative comparison against ViewFormer for reference, we exclude it from numerical comparisons because of its reliance on object masks.
Fig.~\ref{fig:all_colmap} demonstrates the degree of geometric consistency that is attainable by our approach. Tab.~\ref{Tab:CO3D} additionally provides a quantitative comparison against PixelNeRF. On complex scenes rife with ambiguity, the generative nature of our approach enables synthesis of plausible realizations.


\begin{table}[t!]
\centering
\resizebox{\columnwidth}{!}{%
\begin{tabular}{@{}lllllll@{}}
\hline
& & KID$\downarrow$ $\!\!\!\!$& $\!\!$LPIPS$\downarrow$ $\!\!\!\!$&$\!\!\!$ DISTS$\downarrow$ $\!\!\!\!$ & $\!\!\!\!$ PSNR$\uparrow$ $\!\!\!\!\!\!$ & $\!\!$SSIM$\uparrow$ \\
\hline
\multicolumn{2}{l}{$\!\!\!\!$PixelNeRF~\cite{yu2021pixelnerf}}
& 0.210 & 0.705 & 0.487 & 16.26 & 0.271\\
\hline
 \parbox[t]{0mm}{\multirow{2}{*}{\rotatebox[origin=c]{90}{Ours}}}
& $\!\!$ One-Step      & 0.106 & 0.641 & 0.492 & \textbf{16.78} & \textbf{0.331}\\
& $\!\!$ Full          & \textbf{0.012} & \textbf{0.369} & \textbf{0.446} & 15.48 & 0.266\\ 
\hline
\end{tabular}
}
\vspace{3pt}
\caption{  Quantitative comparison of single-view novel view synthesis on CO3D~\cite{reizenstein2021common}.
}\label{Tab:CO3D}
\vspace{-5pt}
\end{table}

\subsection{Ablation Studies}
\label{sec:ablation}

\paragraph{Choice of intermediate representations.}
Tab.~\ref{Tab:ShapeNet} (bottom) compares several choices of intermediate representations within our method. While we have described a specific approach to the task of generative novel view synthesis using diffusion, there is ample freedom to choose how $\denoiser$ interprets information from input views. In fact, the simplest approach forgoes any geometry priors and instead directly conditions the model on an input view by concatenation. In our experiments, this \textit{geometry-free} approach struggled compared to variants that incorporated geometry priors. However, greater model capacity and effective use of cross-attention~\cite{watson2022novel} may be key to making this approach work. We additionally compare against an ``Explicit" intermediate representation similar to our described approach but without the MLP decoder; while slightly faster, this representation generally produced worse results. We compare to the \textit{one-step} inference mode of our method on ShapeNet in Fig.~\ref{fig:shapenet_qualitative} and Tab.~\ref{Tab:ShapeNet}, on MP3D in Fig.~\ref{fig:reb_pixelnerf_mp3d}, and on CO3D in Tab.~\ref{Tab:CO3D}. Like regression-based methods, it obtains excellent PSNR and SSIM scores but lacks the ability to generate plausible results far from the input.
On Matterport3D, Fig.~{\ref{fig:reb_pixelnerf_mp3d}} illustrates the motivation of using a generative prior for long-range synthesis. While the quality of regression-based predictions rapidly degrades with increasing ambiguity, a generative model can create a plausible rendering even in regions with little or no conditioning information, such as behind an occlusion.

\begin{figure}
    \centering
    \includegraphics[width=\linewidth]{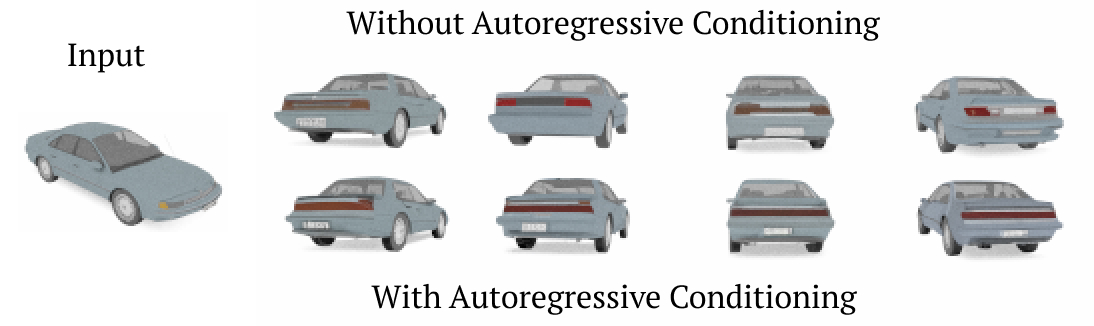}
    \caption{
    \textit{Without} autoregressive conditioning (top), our method generates plausible, albeit geometrically incoherent, novel views conditioned on the input image. \textit{With} autoregressive conditioning (bottom), our method generates plausible sequences that achieve greater geometric consistency between frames.}
    \label{fig:autoregressive_synthesis}
    \vspace{-5pt}
\end{figure}

\paragraph{Effect of autoregressive generation.}
Although autoregressive conditioning slightly trades off image quality (Tab.~\ref{Tab:ShapeNet}), Fig.~\ref{fig:autoregressive_synthesis} demonstrates the necessity of autoregressive conditioning for generating geometrically consistent multi-view images. Without autoregressive conditioning, independently sampled frames are each plausible, but lack coherence—when conditioning information is ambiguous, e.g., when the model is predicting novel views far from the input view, it samples from a wide conditional distribution and accordingly, subsequent frames exhibit significant variance. Autoregressive conditioning effectively conditions the network not only on the source image, but also on previously generated frames that closely overlap with the current view, helping narrow this conditional distribution.

\paragraph{Additional studies.}
Additional ablations, including experiments that evaluate out-of-distribution extrapolation, classifier-free guidance, effect of number of input views, stochastic conditioning, and effect of distance to input views, can be found in the supplement.

\section{Discussion}
\paragraph{Conclusion.}
We proposed a generative novel view synthesis approach from a single image using geometry-based priors and diffusion models. Our hybrid method combines the benefit of explicit 3D representations with the generative power of diffusion models for generating realistic and 3D-aware novel views, demonstrating the state-of-the-art performance in both object-scale and room-scale scenes. We also demonstrate the compelling results on a challenging real-world dataset of CO3D with background\,---\,a challenge never attempted. While our results are not perfect, we believe we presented a significant step towards a practical NVS solution that can operate on a wide range of real-world data.  

\paragraph{Limitations and future work.}
While our method effectively combines explicit geometry priors with 2D diffusion models, the output resolution is currently limited to $128^2$ and the diffusion-based sampling is not fast enough for interactive visualization. Since our model can leverage existing 2D diffusion architectures for $\unet$, it can directly benefit from future advances in the underlying 2D diffusion models. While our method achieves reasonable geometrical consistency, it can still exhibit minor inconsistencies and drift in challenging real-world datasets, which should be addressed by future work. While our method can operate for novel view synthesis from a single view during inference, training the method requires multi-view supervision with accurate camera poses. In this work, we implemented our method using a 3D feature volume representation. Possible future work includes investigating other types of intermediate 3D representations. 

\paragraph{Ethical considerations.}
Diffusion models could be extended to generate DeepFakes. These pose a societal threat, and we do not condone using our work to generate fake images or videos with the intent of spreading misinformation.

\section*{Acknowledgements}
\vspace{-6pt}
We thank David Luebke, Samuli Laine, Tsung-Yi Lin, and Jaakko Lehtinen for feedback on drafts and early discussions. We thank Jonáš Kulhánek and Xuanchi Ren for thoughtful communications and for providing results and data for comparisons. We thank Trevor Chan for help with figures. Koki Nagano and Eric Chan were partially supported by DARPA’s Semantic Forensics (SemaFor) contract (HR0011-20-3-0005). JJ park was supported by ARL grant W911NF-21-2-0104. This project was in part supported by Samsung, the Stanford Institute for Human-Centered AI (HAI), and a PECASE from the ARO. The views and conclusions contained in this document are those of the authors and should not be interpreted as representing the official policies, either expressed or implied, of the U.S. Government. Distribution Statement ``A'' (Approved for Public Release, Distribution Unlimited).

{\small
\bibliographystyle{ieee_fullname}
\bibliography{main}
}

\newpage
\appendix

\section*{Supplementary Material}
\noindent In this supplement, we first provide additional experiments (Sec.~\ref{sec:supp_experiments}). We follow with details of our implementation (Sec.~\ref{sec:implementation}), including further descriptions of the model architecture and training process, as well as hyperparameters. We then discuss experimental details (Sec.~\ref{sec:experiment_details}). Lastly, we consider artifacts and limitations (Sec.~\ref{sec:discussion}) that may be targets for future work. We encourage readers to view the accompanying supplemental videos, which contain additional visual results.

\section{Additional experiments \& ablations}
\label{sec:supp_experiments}


\begin{figure}[h]
    \centering
    \includegraphics[width=\linewidth]{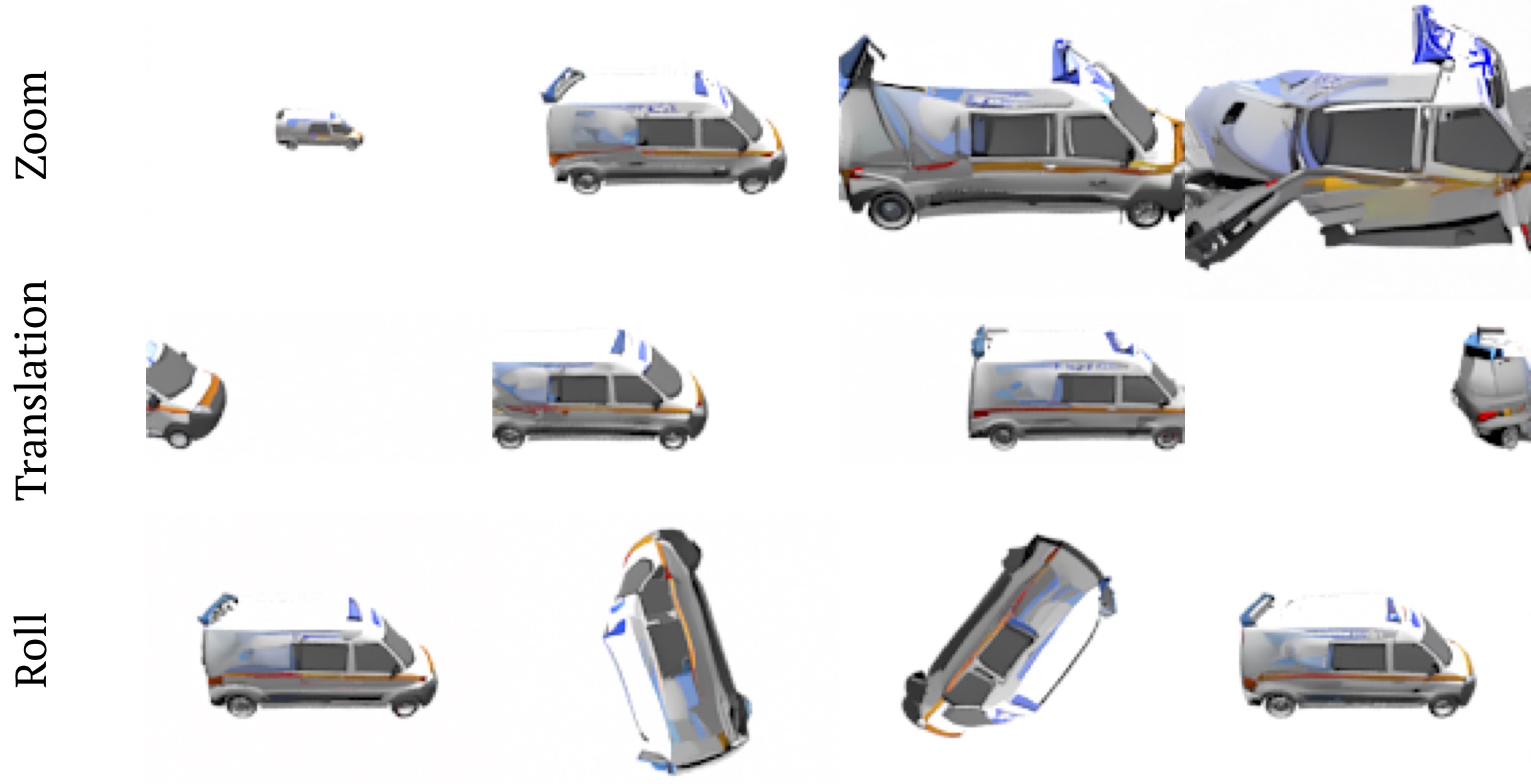}
    \caption{New views generated from out-of-distribution poses. Extreme zooms and large translations may lead to unrealistic views.}
    \label{fig:out-of-distrib}
\end{figure}

\subsection{Extrapolation to unseen camera poses.}

In the ShapeNet dataset, cameras are located on a sphere, point towards the centers of the objects and have the same ``up'' direction during training. We investigate the results of our method when querying out-of-distribution poses at test time in Fig.~\ref{fig:out-of-distrib}. From a fixed pose, we generate a zoom, a one-dimensional translation of the camera, and a camera roll. Although novel views deteriorate with large deviations from the training pose distribution, the 3D prior present in our method can reasonably tolerate small extrapolations.

\subsection{Percentile results based on LPIPS}
\begin{figure}
    \centering
    \includegraphics[width=\linewidth]{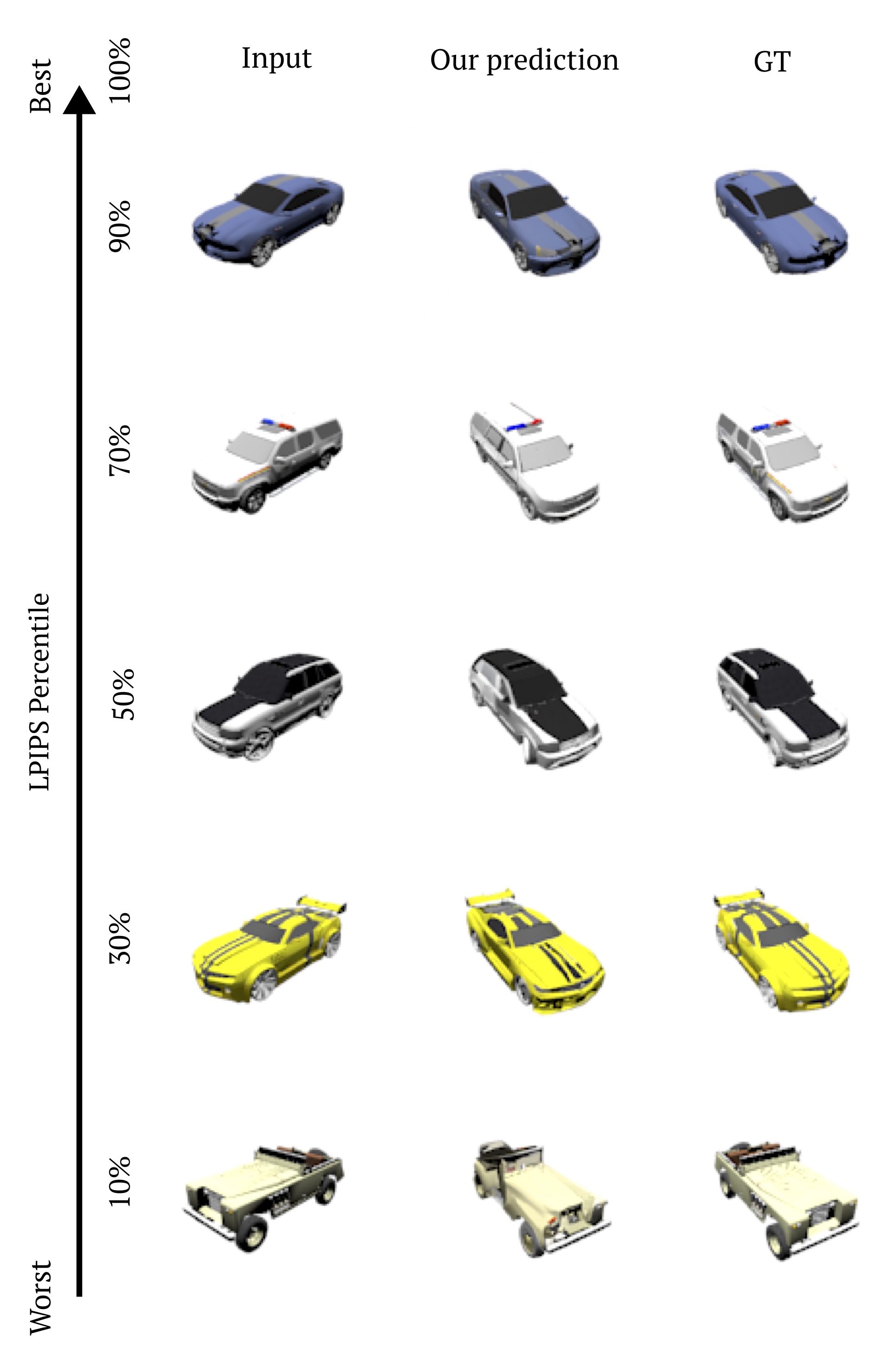}
    \caption{Our synthesized novel views sorted by the percentile of the LPIPS~\cite{zhang2018perceptual} score, with results that scored best according to LPIPS at the top.}
    \label{fig:LPIPS}
\end{figure}

Fig.~\ref{fig:LPIPS} shows our synthesized results on ShapeNet ordered by the percentile of the LPIPS~\cite{zhang2018perceptual} score, with examples that scored best according to the metric at the top and examples that scored worst at the bottom. We compute predictions for the same input and output views across the entire test set. To reduce the effects of randomness, we evaluate 9 realizations for each input, and use only the median image/score when ordering our results. Our method produces consistently sharp outputs (even at the 10\textsuperscript{th} percentile) and maintains overall textures and shapes from the input image.  


\subsection{Handling multiple input images}

\begin{figure}
    \centering
    \includegraphics[width=\linewidth]{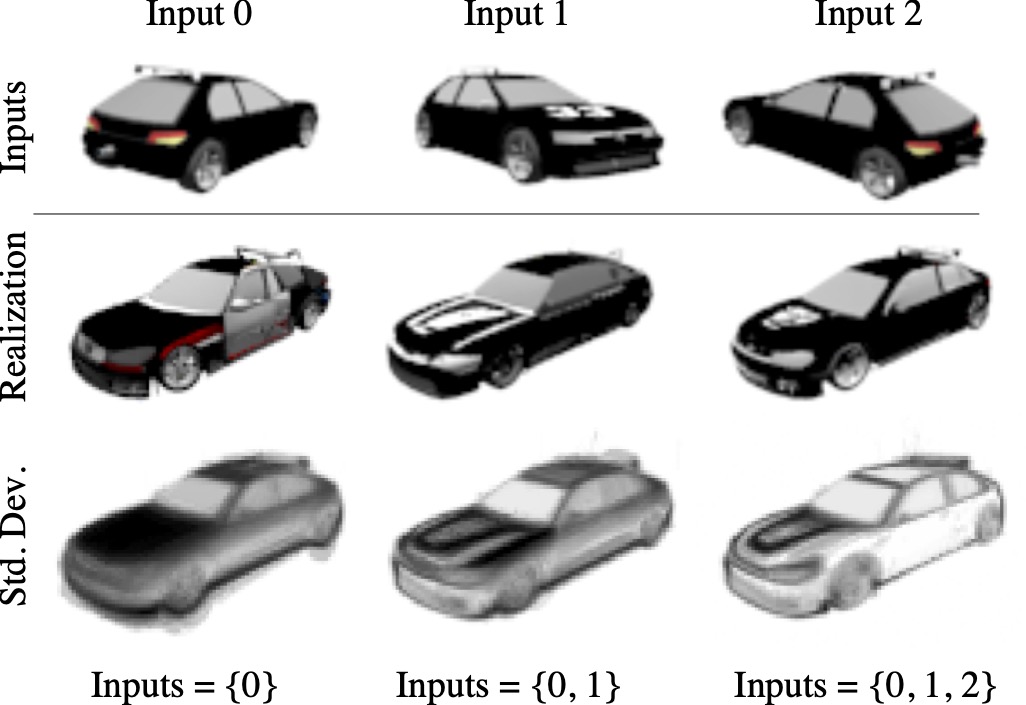}
    \caption{Effect of varying the number of input views. Increasing the number of input views reduces uncertainty, decreasing the pixel-wise standard deviation in novel renderings. Dark pixels in the third row represent higher standard deviation and indicate greater variation in the realizations.}
    \label{fig:n_inputs}
\end{figure}

Fig.~\ref{fig:n_inputs} shows our generated novel views when more than one image is given as the input conditioning information. 
When only 1 view is given from the back side of the car, the model has the freedom to choose multiple plausible completions for the unseen front side of the car, leading to a high standard deviation (high uncertainty). Adding 2 or 3 views reduces uncertainty (low standard deviation), and the model generates a novel view that is compatible with multiple input views.

\subsection{Effect of distance to input view}
\begin{figure}
    \centering
    \includegraphics[width=\linewidth]{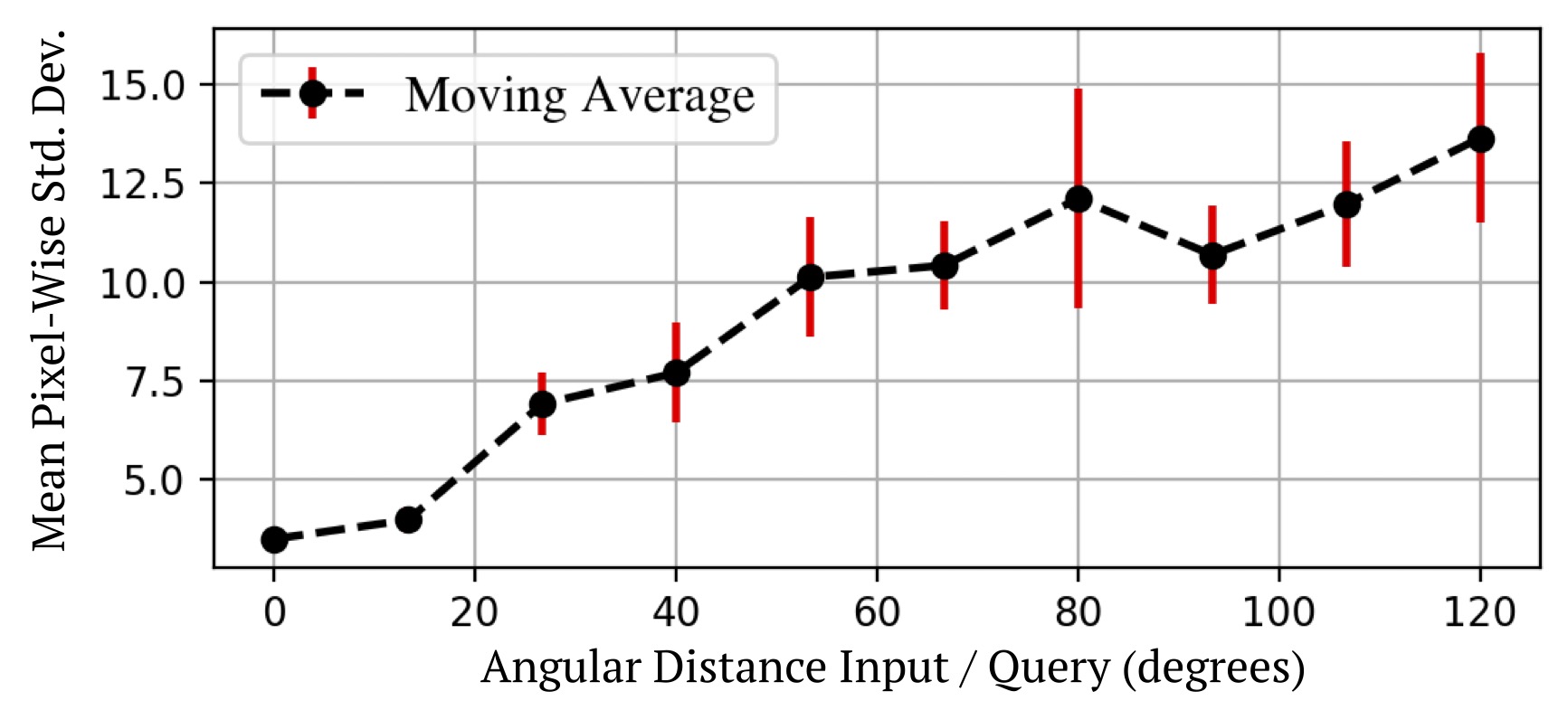}
    \caption{Average pixel variance of generated views vs. the distance between the query camera and the input camera. Input views close to the camera are valuable—the model can directly observe many of the details it must transfer to the output rendering. Input views distant to the camera are more ambiguous—the model is tasked with generating large parts of the rendering from scratch. As the conditioning information gets increasingly ambiguous, novel views get increasingly diverse. Pixel variance is calculated across 50 renderings per pose. Red bars indicate the empirical standard deviation of the moving average.}
    \label{fig:std-vs-angle}
\end{figure}

As Fig.~\ref{fig:std-vs-angle} demonstrates, nearby views provide more valuable information than distant views, thus reducing variance in the output rendering. Consequently, by conditioning autoregressively on nearby views, we narrow the conditional distribution of possible outputs, improving geometric consistency compared to non-autoregressive conditioning.






\subsection{Classifier-free guidance}

Recently, \cite{ho2022classifier} suggested classifier-free diffusion guidance technique to effectively trade off diversity and sample quality. At training, we implement classifier-free guidance by dropping out the feature image with 10\% probability; in its place, we replace this conditioning image with a sample of Gaussian noise. At inference, we can linearly interpolate between unconditional and unconditional predictions of the denoised image in order to boost or decrease the effect of the conditioning information.


\begin{figure}
    \includegraphics[width=\linewidth]{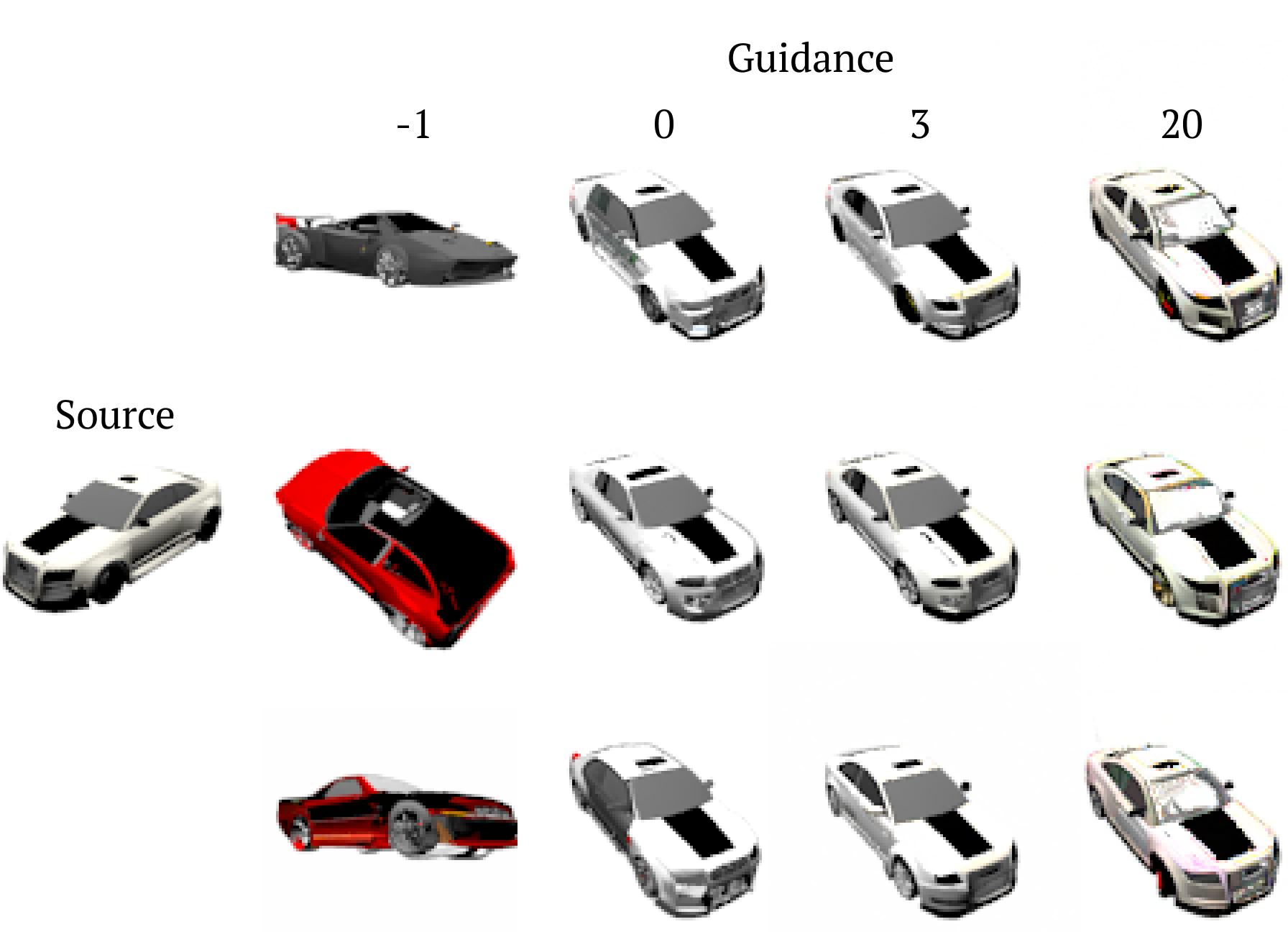}
    \caption{Independent (single-frame input) NVS with various classifier-free guidance (CFG) strengths. For each level of CFG, we show three realizations. With $\text{guidance}=0$, we sample a ``diverse" set of novel views, each plausible, but with variations (e.g. doors). Higher guidance strength reduces diversity but improves sample quality. Excessively high guidance begins to introduce saturation and visual artifacts. Negative guidance upweights the unconditional contribution; with $\text{guidance}=-1$, generation is unconditional.}
    \label{fig:supp_classifier-free-guidance-independent.jpg}
\end{figure}
\begin{figure}
    \includegraphics[width=\linewidth]{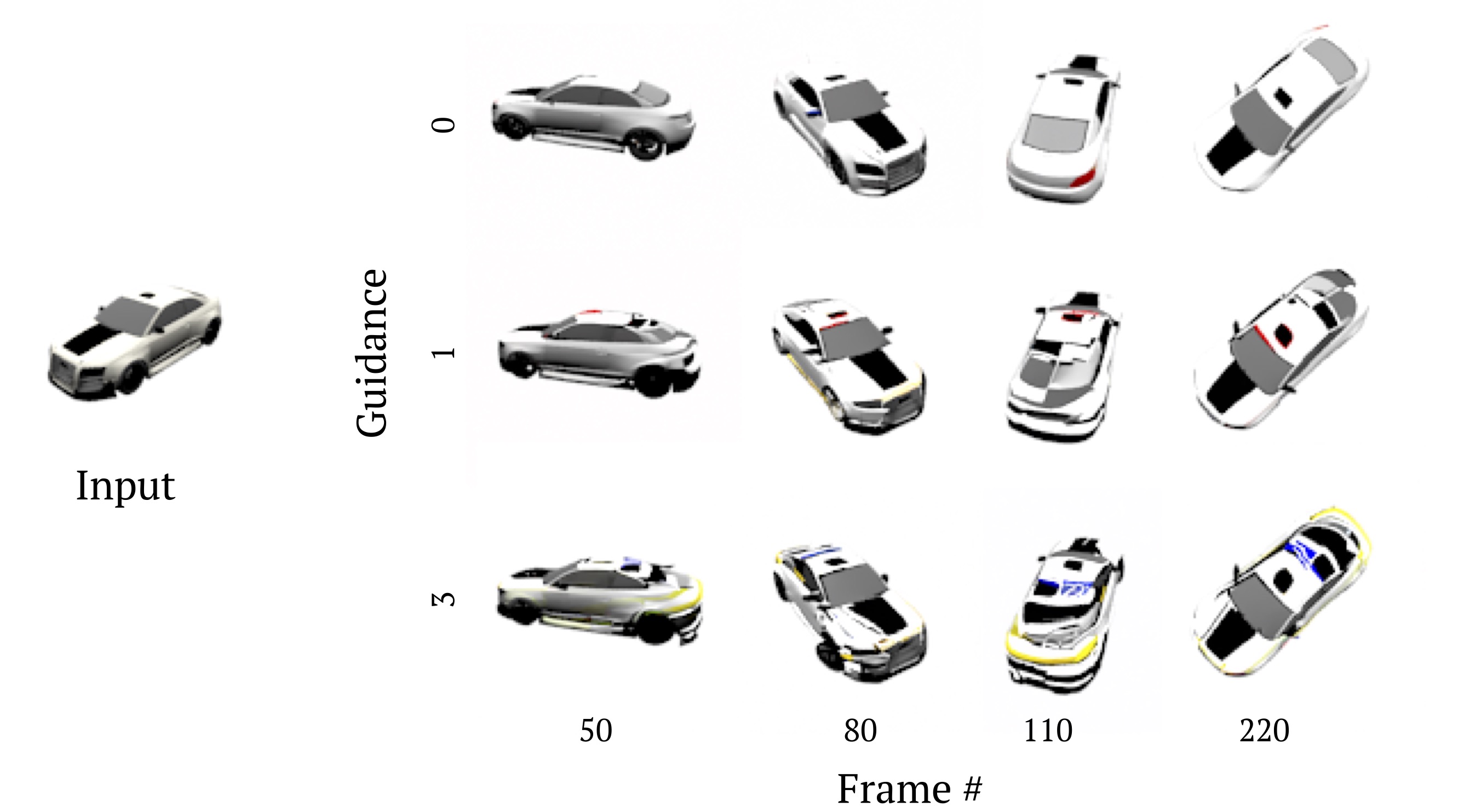}
    \caption{ Autoregressive sequence generation with varying CFG strength. With low guidance, we can generate extended autoregressive sequences with little deterioration over time. Higher guidance tends to carry over errors from previous frames, which gradually degrades the quality of subsequent generations.}
    \label{fig:supp_classifier-free-guidance-autoregressive}
\end{figure}


Fig.~\ref{fig:supp_classifier-free-guidance-independent} shows the effect of classifier-free guidance~\cite{ho2022classifier} (CFG) when making predictions in isolation. In general, positive classifier-free guidance increases the effect of the conditioning information and improves sample quality. With $\text{guidance}=0$, our model produces greater variation of generated views (note the different realizations of the passenger-side door). However, we would consider some of these realizations to be unlikely given the input. Increased CFG strength narrows the distribution of possible outputs, and while we would consider such a set of realizations to be less diverse, each one is of high fidelity. Excessively high guidance strength begins to introduce artifacts and color saturation. Negative guidance upweights unconditional prediction; $\text{guidance}=-1$ produces unconditional samples without influence from the input image. In general, when making independent novel view predictions, we find moderate levels of CFG to be beneficial. However, as described in Sec.~\ref{supp_extended_autoregressive_generation}, CFG has an adverse effect on the quality of autoregressively generated sequences. As a default, we refrain from using CFG in our experiments.

\subsection{Extended autoregressive generation}
\label{supp_extended_autoregressive_generation}
Fig.~\ref{fig:supp_classifier-free-guidance-autoregressive} shows autoregressively generated sequences made with varying levels of  classifier-free guidance. When making long autoregressive sequences, the ability to suppress errors and return to the image manifold is an important attribute. Unchecked, gradual accumulation of errors could lead to progressive deterioration in image quality. Intuitively, unconditional samples do not suffer from error buildup, since unconditional ($\text{CFG}=-1$) samples make use of \emph{no} information from previous frames. On the other end of the spectrum, highly conditioned ($\text{CFG}>>0$) samples should be more likely to suffer from error accumulation because they \emph{emphasize} information from previous frames. A happy medium between these two extremes allows the model to use information from previous frames while preventing undesired error accumulation. Empirically, we find that while small positive guidance can reduce frame-to-frame flicker, it enhances the model's tendency to carry over visual errors from previous frames. We observe saturation buildup and artifact accumulation to be significant roadblocks to using CFG when synthesizing long video sequences. For these reasons, we default to using $\text{CFG}=0$, which we found to enable autoregressive generation of long sequences without significant error accumulation. A solution that enables higher CFG weights for autoregressive generation may make a valuable contribution in the future. 

\subsection{Alternative autoregressive conditioning schemes}
\label{sec:supp_alternative_autoregressive_strategies}

\paragraph{Baseline strategy}

When generating a sequence autoregressively, there are many possible strategies, each with a set of tradeoffs. To produce the visual results presented in our work, we used the following baseline strategy, with minor variations for different datasets. As described in the main paper, our baseline strategy is to condition our model on the input image(s), the most recently generated rendering, and five previously generated images, selected at random.

For Matterport3D, when generating long sequences, we select the five previously generated frames from a set of only the 20 most recently generated frames; we additionally condition on every 15\textsuperscript{th} previously generated frame.

For CO3D, we use the two-pass conditioning method discussed below to improve temporal consistency.





\begin{figure*}
    \centering
    \includegraphics[width=\textwidth]{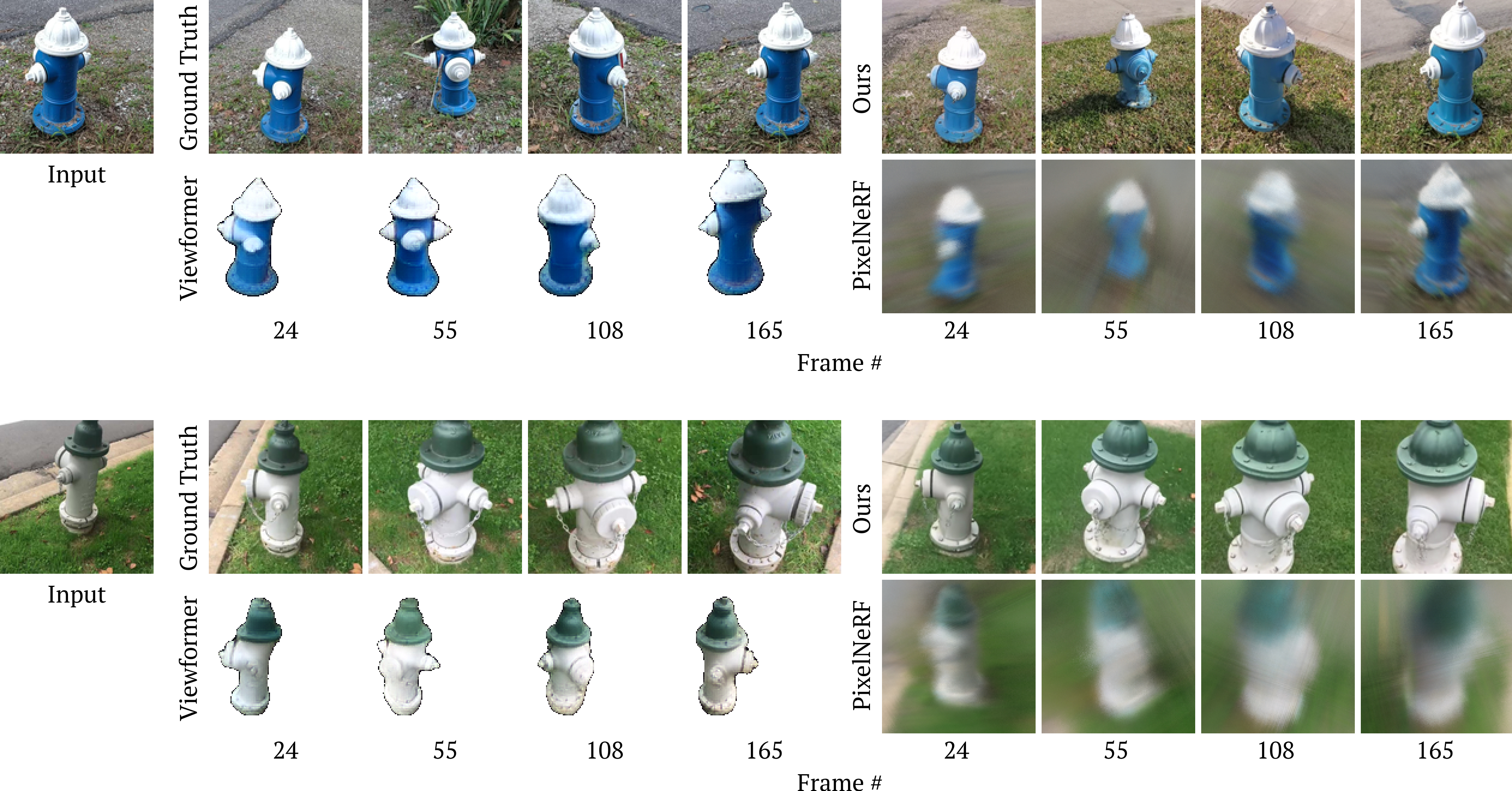}
    \caption{Qualitative comparison for single-view novel view synthesis on CO3D~\cite{reizenstein2021common} \emph{Hydrants}.}
    \label{fig:supp_co3d_qualitative}
\end{figure*}
 \begin{figure*}
    \centering
    \includegraphics[width=\textwidth]{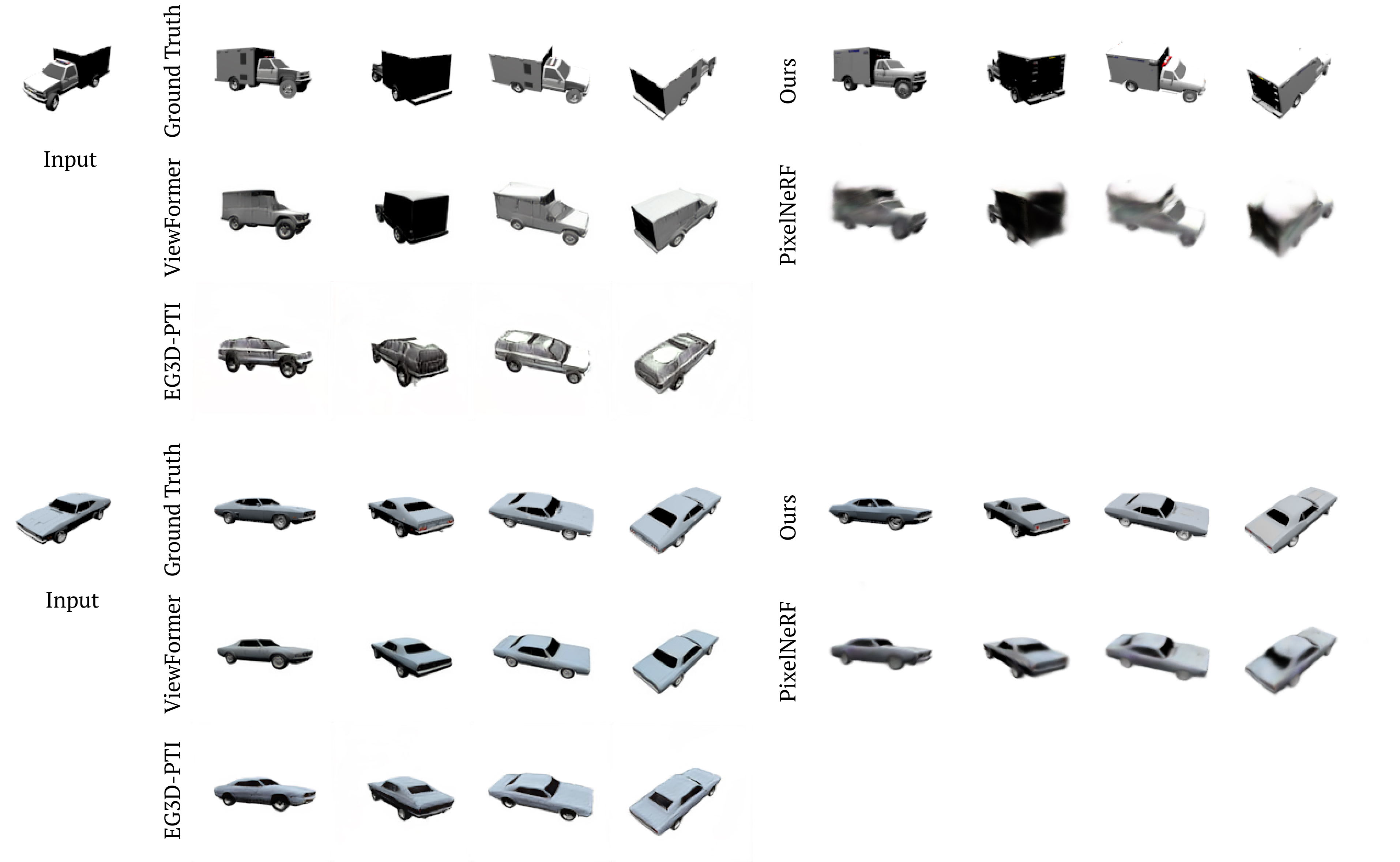}
    \caption{Additional qualitative comparisons against baselines on ShapeNet~\cite{chang2015shapenet}.}
    \label{fig:shapenet_additional_comarison}
\end{figure*}

\paragraph{Alternative strategies and tradeoffs}
As described in the main paper, our baseline autoregressive strategy can induce noticeable flickering. One way to reduce flickering is to condition on \textit{only} the previous frame. Doing so almost completely eliminates frame-to-frame flicker. However, this strategy sacrifices long-term consistency and does little to prevent drift; new renderings might not be consistent with frames rendered at the start of the sequence. By contrast, to promote long-term consistency, one could avoid conditioning on previously-generated frames at all and instead condition on only the input image(s). Because drift is the result of error accumulation from conditioning on previous generations, this strategy eliminates potential for drift. However, it suffers from short-term inconsistency (i.e. frame-to-frame flicker). 
We found our baseline strategy, which conditions on the inputs, the most recent rendering, and several previous renderings, to be a good compromise between long-term and short-term consistency. The number of previously generated images we condition upon affects the behavior. Because we equally weight the contribution of all images we condition upon, increasing the number of previous renderings (which are sampled uniformly from the generated sequence) reduces the relative contribution of the most recent rendering. Increasing the size of this ``buffer'' of previously-generated conditioning images thus improves long-term consistency at the cost of short-term consistency; reducing the size of the buffer has the opposite effect.

One way to suppress flickering is to generate frames in two passes, where in the second pass, we condition on the nearby frames from the first pass in a sliding window fashion. Empirically, conditioning on only the nearest 4 frames during the second pass results in videos with reduced flicker, at the expensive of higher inference computation. However, unless otherwise noted, we render all videos shown with our baseline autoregressive strategy, i.e. \textit{without} these alternative methods. 

\subsection{Stochastic Conditioning}
\begin{figure}
    \centering
    \includegraphics[width=\linewidth]{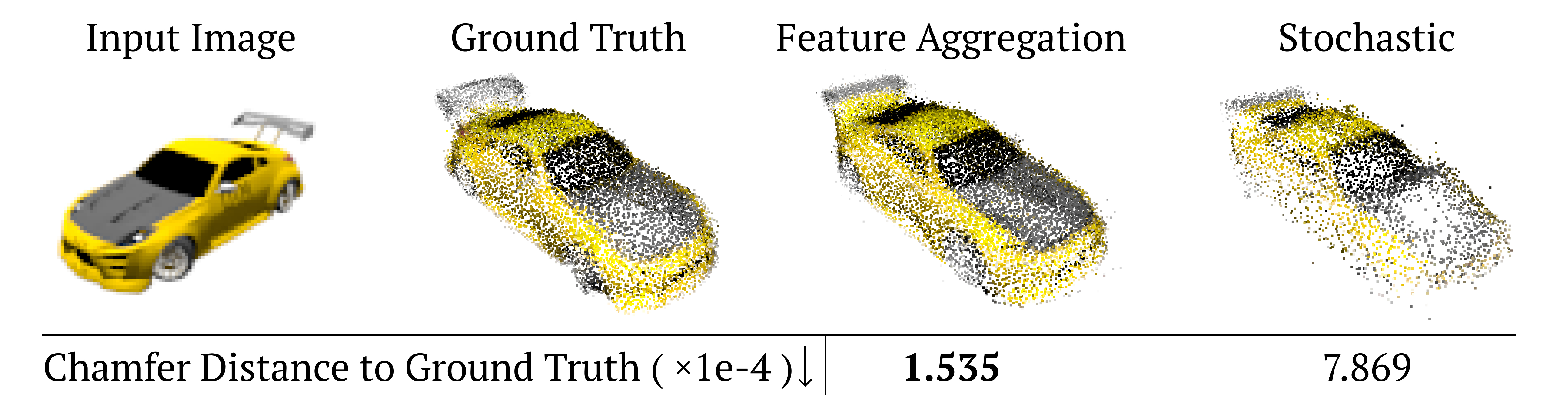}
    \caption{Our default autoregressive conditioning strategy, which aggregates information from multiple views within a feature volume, typically performs at least on par with stochastic view conditioning~\cite{watson2022novel} in geometric consistency, but requires many fewer steps of diffusion to remain effective. Here, we compare COLMAP reconstructions of a sequenced produced by feature aggregation, using 25 steps of denoising, against a sequence produced by stochastic conditioning, using 256 steps of denoising.}
    \label{fig:stochastic_conditioning}
\end{figure}
%
To demonstrate the effectiveness of our autoregressive synthesis method, which aggregates conditioning feature volumes from autoregressively selected generated images, we compare to an adaptation of the stochastic conditioning method proposed in 3DiM~\cite{watson2022novel}.
We adapt the stochastic conditioning method to our architecture by replacing the feature volume aggregation from autoregressively selected generated images with a single feature volume generated from an image randomly sampled from all previously generated images. As done in 3DiM, the number of diffusion denoising steps is increased significantly and the randomly sampled image is varied at each individual step of denoising. Each generated final image is then added to the set of all previous images and can be used as conditioning in subsequent view generations.
This alternative form of conditioning is also able to provide the model with information from many generated views, but they are processed independently with each step of denoising, rather than together after a feature volume aggregation.

In Fig.~\ref{fig:stochastic_conditioning}, we show 3D reconstruction results from sequences of images generated by our autoregressive synthesis method and with our adaptation of stochastic conditioning~\cite{watson2022novel}.
Here, we find that our autoregressive synthesis method performs slightly better than stochastic conditioning in terms of 3D consistency of generated frames as seen by the COLMAP 3D reconstruction and corresponding Chamfer distance. Additionally, we are able to generate novel views significantly faster -- in practice, stochastic conditioning requires $256$ denoising steps to generate each novel view while our method only requires $25$, leading to a $10x$ improvement in speed.





\subsection{Additional Common Objects in 3D results}

We provide additional results for single-view novel view synthesis (NVS) with real-world objects for CO3D \textit{Hydrants} in Fig.~\ref{fig:supp_co3d_qualitative}. We compare against ViewFormer~\cite{kulhanek2022viewformer}, which has demonstrated success in few-shot NVS on CO3D, and PixelNeRF~\cite{yu2021pixelnerf}. However, we note that ViewFormer is not a 1:1 comparison for two reasons: 1. ViewFormer operates with object masks, whereas our method operates with backgrounds. 2. ViewFormer train/test splits did not align with other methods. For this figure, and for comparison videos, we selected objects that were contained in our \emph{test} split but were part of ViewFormer's \emph{train} split. Despite these disadvantages, our method demonstrates a compelling ability to plausibly complete complex scenes.



\subsection{Additional ShapeNet results}

Fig.~\ref{fig:shapenet_additional_comarison} provides additional visual comparisons on the ShapeNet~\cite{chang2015shapenet} dataset against baselines. In general, our method renders images with sharper details and higher perceived quality than PixelNeRF, while better transferring details from the input image than ViewFormer and EG3D. In this figure, renderings from our method are selected from autoregressively-generated sequences.

\section{Implementation details}
\label{sec:implementation}

We implemented our 3D-aware diffusion models using the official source code of EDM~\cite{karras2022elucidating}, which is available at \url{https://github.com/NVlabs/edm}. Most of our training setup and hyperparameters follow~\cite{karras2022elucidating}; the exceptions are detailed here.

\paragraph{Feature volume encoder, $T$.}
Our encoder backbone is based on DeepLabV3+~\cite{chen2018encoder}. We use a Pytorch reimplementation~\cite{Iakubovskii:2019} available at \url{https://github.com/qubvel/segmentation_models.pytorch}, and ResNet34~\cite{he2016deep} as the encoder backbone. We found unmodified DeepLabV3+, to struggle because the output branch contains several unlearned, bilinear upsampling layers; this resolution bottleneck makes it difficult to effectively reconstruct fine details from the input. We replace these unlearned upsampling layers with learnable convolutional layers and skip connections from previous layers. We disable batchnorm and dropout throughout the feature volume encoder. The feature volume encoder expects as input a $3\times128\times128$ image; it produces a $(16\times64)\times128\times128$ feature image, which we reshape into a $16\times64\times128\times128$ volume.

\paragraph{Multiview aggregation.}
We aggregate information from multiple input views by predicting a feature volume $\boldsymbol{W}_i$ for each input image independently, projecting the query point into each feature volume, sampling a separate feature vector from each feature volume, and mean-pooling across the sampled feature vectors to produce a single aggregated feature. We experimented with two alternative aggregation strategies: 1. max-pooling, and 2. weighted average pooling, where the feature volumes have an additional channel that is interpreted as a weight by a softmax function. We found these alternative aggregation strategies to perform similarly to mean-pooling.

\paragraph{MLP, $f$.}
We use a two-layer ReLU MLP to aggregate features drawn from multiple input images. Our MLP has an input dimension of $16$, two hidden layers of dimension $64$, and an output dimension of $17$, which is interpreted as a 1-channel density $\tau$ and a 16-channel feature $\boldsymbol{c}$. We additionally skip the MLP's input feature to the output feature.

\paragraph{Rendering.}

We render feature images from the model using neural volume rendering~\cite{mildenhall2021nerf} of features~\cite{niemeyer2021giraffe}, from the neural field parameterized by the set of feature volumes $\boldsymbol{W}$ and the MLP $f$. For computational efficiency, we render at half spatial resolution, i.e. $64 \times 64$ and use bilinear upsampling to produce a $128 \times 128$ feature image. We use 64 depth samples by default, scattered along each ray with stratified sampling. We do not use importance sampling.

\paragraph{UNet, $U$.}
The design of $U$ is based on \emph{DDPM++}~\cite{song2020score}, using the implementation and preconditioning scheme of \cite{karras2022elucidating}. $U$ accepts as input 19 total channels (a noisy RGB image, plus a 16-channels feature rendering) of spatial dimension $128^2$. It produces a 3-channel $128^2$ denoised rendering. For experiments shown in the manuscript, our models contain five downsampling blocks with channel multipliers of [128, 128, 256, 256, 256]. As in \cite{song2020score}, we utilize a residual skip connection from the input to $U$ to each block in the encoder of $U$. 

\paragraph{Training.}

We use a batch size of 96 for all training runs, split across 8 A100 GPUs, with a learning rate of $2 \times 10^{-5}$. During training, we sample the noise level $\sigma$ according to the method proposed by \cite{karras2022elucidating} by drawing $\sigma$ from the following distribution: 

\begin{equation}
    \log(\sigma) \sim \mathcal{N} (P_\text{mean}, P^2_\text{std}).
\end{equation}

We use $P_\text{mean} = -1.0$, $P_\text{std} = 1.4$.
During training, we randomly drop out the conditioning information with a probability $0.1$ to enable classifier-free guidance. In place of the rendered feature image, we insert random noise.

Our dataset is composed of posed multi-view images, where for each training image, we are given the $4 \times 4$ camera pose matrix, the camera field of view, and a near/far plane. For all experiments, we specify a global near/far value for each dataset, where the values are chosen such that a camera frustum with the chosen near/far planes adequately covers the visible portion of the scene. For ShapeNet, $\text{near/far} = (0.8, 1.8)$; for MP3D, $\text{near/far} = (0., 12.5)$; for CO3D, $\text{near/far} = (0.5, 40)$. We found our method to be fairly robust to the chosen values of near/far planes.

For ShapeNet, we train until the model has processed 140M images, which takes approximately 9 days on eight A100 GPUs. For MP3D, we train for 110M images, which takes approximately 7 days on eight A100 GPUs. For CO3D, we train for 170M images, which takes approximately eleven days on eight A100 GPUs.

\paragraph{Augmentation.}

During training, we introduce two forms of augmentation. First, with probability $0.5$, we add Gaussian white noise to the input images. For input images in the range $[-1, 1]$, we sample the standard deviation of the added noise uniformly from $[0, 0.5]$. Second, we apply non-leaking augmentation~\cite{karras2022elucidating} to $U$. With probability 0.1, we apply random flips, random integer translations (up to 16 pixels), and random 90º rotations, where the transformations are applied to the input noisy image, the input feature image, and the target denoised image. We condition $U$ with a vector that informs it of the currently applied augmentations; we zero this vector at inference.

\paragraph{Inference.}

We use the deterministic second order sampler proposed in \cite{karras2022elucidating} at inference. As a default, we use $N=25$ timesteps, with a noise schedule governed by $\sigma_\text{max} = 80$, $\sigma_\text{min} = 0.002$, and $\rho = 7$, where $\rho$ is a constant that controls the spacing of noise noise levels. The noise level at a timestep $i$ is given in Eq.~\ref{eq:supp_noise_levels}:

\begin{equation} \label{eq:supp_noise_levels}
    \sigma_{i < N} = \left( {\sigma_\text{max}}^\frac{1}{\rho} + \frac{i}{N-1}\left({\sigma_\text{min}}^\frac{1}{\rho} - {\sigma_\text{max}}^\frac{1}{\rho}\right)\right).
\end{equation}

Rendering an image from scratch with 25 denoising steps takes approximately $1.8$~seconds per image at inference on an RTX 3090 GPU.



\paragraph{``Production" settings for CO3D.}
For rendering videos of CO3D, we use more computationally expensive ``production'' hyperparameters to obtain better image quality. Seeking better image quality and detail, we use 256 denoising steps instead of the default 25 denoising steps. Seeking better temporal consistency, we increase the number of samples per ray cast through the latent feature field, from 64 to 128; we also use the two-pass form of autoregressive conditioning described in Sec.~\ref{sec:supp_alternative_autoregressive_strategies}.


\section{Experiment details}
\label{sec:experiment_details}

\subsection{Evaluation details}
\paragraph{FID Calculation.} We compute FID by sampling 30,000 images randomly from both the ground truth testing dataset and corresponding generated frames. We use an inception network provided in the StyleGAN3~\cite{Karras2021} repository for computing image features.

\paragraph{KID Calculation.} We compute KID by sampling all images from both the ground truth testing dataset and corresponding generated frames. We use the implementation of {\em clean-fid} \cite{parmar2021cleanfid}, available at \url{https://github.com/GaParmar/clean-fid}.

\paragraph{COLMAP Reconstructions.} We compute COLMAP reconstructions using frames from rendered video sequences. We provide the ground-truth camera pose trajectory as input for all reconstructions. For ShapeNet evaluations, we additionally compute masks by thresholding images to remove white pixels. We leave all settings at their recommended default.

\paragraph{Chamfer Distance Calculation.}
For all datasets, we compute the bi-directional Chamfer distance between the reconstructed point cloud from synthesized images to the reconstructed point cloud from ground truth images. Additionally, for CO3D, we translate and scale the reconstructed point clouds to lie within the unit cube.

\subsection{Baselines}
\paragraph{PixelNeRF~\cite{yu2021pixelnerf}.} We compare to PixelNeRF for the ShapeNet and CO3D single-image novel view synthesis benchmark. For ShapeNet, we use the official implementation and pre-trained weights for single-category (car), single-image, ShapeNet novel view synthesis evaluation provided at: \url{https://github.com/sxyu/pixel-nerf}. We follow the protocol described in the original PixelNeRF paper and SRNs~\cite{sitzmann2019scene} for data pre-processing. We use the provided dataset and splits in the PixelNeRF repository for training and testing of both our method and PixelNeRF (this dataset is slightly different from that used in the SRNs paper due to a bug; see PixelNeRF supplementary information). We follow the same protocol for evaluation as we do for our method and SRNs: view 64 is used as input, and the remaining 249 views are synthesized conditioned on this. For CO3D, we train PixelNeRF from scratch using our train/test splits and using the recommended hyperparameters.

\paragraph{ViewFormer~\cite{kulhanek2022viewformer}.} We compare to ViewFormer on the ShapeNet single-image novel view synthesis benchmark and qualitatively on single-image novel view synthesis for CO3D. We received the data and results for single-image novel view synthesis for the entire ShapeNet testing set from the authors. We compute metrics using their provided ground truth data and synthesized results. The training and testing splits are the same as those used in our method and in PixelNeRF. They use the previously introduced protocol for single-image novel view synthesis evaluation: view 64 is used as input, and the remaining 249 views are synthesized conditioned on this. For CO3D, we instead condition on the first frame from each shown sequence, and generate a video based on this conditioning information. We use provided source code from the official repository at: \url{https://github.com/jkulhanek/viewformer}. We do not generate quantitative metrics, as ViewFormer operates on masked and center-cropped images. Additionally, the images, which we use for comparison are in the training set for ViewFormer, while for our method they are in the test set.

\paragraph{Look Outside the Room~\cite{ren2022look}.} We compare against Look-outside-the-room (LOTR), the current state-of-the-art method on novel view synthesis on Matterport3D (MP3D)\cite{chang2017matterport3d} and RealEstate10K \cite{zhou2018stereo} datasets. For LOTR, we obtained the pretrained weights for the MP3D dataset from their official codebase \url{https://github.com/xrenaa/Look-Outside-Room}. We match LOTR's data preparation methodology, including identical train/test splits, and we use LOTR's implementation for generating multi-view images from MP3D RGB-D scans. For testing their method, we prepare a common set of 200 input images from the test split with the trajectories and ground truth images for the next 10 frames for each input. Then, we run the LOTR method on the given input using the code from their Github repository, using 3 overlapping frame windows, as stated in their paper. We run LOTR on the next 10 frames, given the input frame, and measure the metrics against the ground truth. 

\paragraph{Additional Baselines for MP3D} To further evaluate our method's effectiveness on the novel-view synthesis task on MP3D scenes, we compare against additional baselines of GeoGPT \cite{rombach2021geometryfree} and SynSin \cite{wiles2020synsin}. Note that these two baselines, along with another recent work of PixelSynth \cite{rockwell2021pixelsynth}, have been already shown to underperform against LOTR \cite{ren2022look}. Since GeoGPT does not provide pre-trained models or rendered images for MP3D, we inquired the authors of LOTR for the images they used for the benchmarks. The acquired NVS images of GeoGPT and SynSin are rendered by the exact same protocol as our experiments, except that they proceeded five frames from the initial input images for 200 sequences (thus we have 1,000 images in total). We note that the trajectories used for these acquired images are different from the trajectories we used for our experiments because the trajectories are generated randomly via the Habitat embodied agent simulation \cite{savva2019habitat}. However, at 1000 trajectory samples, we believe our comparisons are statistically significant. The final numbers we computed show similar trends to those reported in the LOTR paper, further confirming the validity of the comparisons. Both qualitatively and quantitatively, we observe that our novel-view renderings are significantly more desirable.

\subsection{Dataset details}

\paragraph{ShapeNet~\cite{chang2015shapenet}.} We extensively evaluate our method on the ShapeNet dataset. The full ShapeNet dataset contains different object categories, each with a synthetically generated posed images in pre-defined training, validation, and testing sets. In our work, we specifically evaluate with the ``cars" category, and focus on single-image novel view synthesis. We use the version of the dataset provided in PixelNeRF~\cite{yu2021pixelnerf} for consistency in training and evaluation, keeping all frames in the dataset at $128^2$ resolution and doing no additional pre-processing. As described in the main paper, the training set contains 2,458 cars, each with 50 renderings randomly distributed on the surface of a sphere. The test split contains 704 cars, each with 250 rendered images and poses on an Archimedean spiral. During the training of our method, we use the defined training split, randomly sampling between one and three input frames with the objective of synthesizing a randomly selected target frame for a specific object instance. In evaluation, we use the defined testing split, use image number 64 as input, and synthesize the other 249 ground truth images. We note that since these images are synthetically generated at only $128^2$, they lack backgrounds and fine detail. However, the accuracy of poses in the constrained environment and consistent evaluation method between baselines allows for easily providing quantitative benchmarks for single-image novel view synthesis.

\paragraph{Matterport3D~\cite{chang2017matterport3d}.}  We showcase our algorithm on a highly complex, large-scale indoor dataset, Matterport3D (MP3D). MP3D contains RGB-D scans of real-world building interiors. Scenes are calibrated to metric scale, and thus there is no scale ambiguity. We preprocess MP3D scans into a dataset of posed multi-view images following the procedure detailed in LOTR~\cite{ren2022look} and SynSin~\cite{wiles2020synsin}. Specifically, we generate the image sequences by simulating a navigation agent in the room scans, using the popular Habitat \cite{savva2019habitat} API.
We randomly select the start and end position within the MP3D scenes and simulate the navigation towards the goal via Habitat. The agent is only allowed to take limited actions, including going forward and rotating 15 degrees.
During training, we randomly sample a target frame and then select 1 to 3 random source frames in the neighborhood of 20 frames for conditioning.

\paragraph{Common Objects in 3D~\cite{reizenstein2021common}.} We validate our method on a real-world dataset: Common Objects in 3D (CO3D).
The CO3D dataset consists of several categories. We train on CO3D Hydrants, which contains 726 scenes. The average scene consists of around 200 frames of RGB video, object masks, poses, and semi-sparse depth. We note that the CO3D dataset is quite unconstrained: even across scenes within a category, aspect ratio, resolution, FOV, camera trajectory, object scale, and global orientation all vary.
Additionally, we note that the dataset is noisy, with several examples of miscategorized objects and numerous extremely short or low-quality videos. Such noise adds to the challenge of single-image NVS.

In preparing data, we first center-crop to the largest possible square, then resize to $128^2$ using Lanczos resampling. We adjust the camera intrinsics to reflect this change. We also seek to normalize the canonical scale of scenes across the dataset. To do so, we examine the provided depths within each scene, and consider the depth values that fall within the object segmentation mask. For each image, we calculate the median value of the masked depth. Taking the mean of these median values across the scene gives us a rough approximation of the distance between the camera and object. We adjust the scale of the scene so that this camera-object distance is identical across every scene in the dataset.




To help resolve scale, which is highly variable across the dataset, and to provide information parity with PixelNeRF, which has access to a global reference frame, we provide our feature encoder, $T$, with the location of the global origin. In addition to each input RGB image, we concatenate a channel that contains a depth rendering of the three coordinate planes, as rendered from the input camera. We modify $T$ to accept the four-channel input. We find this input augmentation to improve our model's ability to localize objects.

\section{Discussion}
\label{sec:discussion}

\subsection{Alternative approaches}

\paragraph{GAN-based generative novel view synthesis.}

We have presented a diffusion-based generative model for novel view synthesis, but in principle, it is possible to construct a similar framework around other types of generative models. Generative Adversarial Networks~\cite{goodfellow2020generative} (GANs), are a natural fit, and adversarial training could drop in to replace our diffusion objective with minor changes. While recent work~\cite{dhariwal2021diffusion} has demonstrated that diffusion models often outperform GANs in mode coverage and image quality, GANs have a major advantage in speed. Future work that aims for real-time synthesis may prefer a GAN-based 3D-aware NVS approach.




\paragraph{Transformer-based, geometry-free multi-view aggregation strategies.}

A promising alternative to explicit geometry priors, such as the type we have presented in this work, is to instead make use of powerful attention mechanisms for effectively combining multiple observations. Scene Representation Transformers~\cite{sajjadi2022scene} utilize a transformer-based approach to merge information from multiple views, which is effective for NVS on both simple and complex scenes. We explored an SRT-based variant of $\Gamma$, which would forego explicit geometry priors for a transformer and light field~\cite{sitzmann2021lfns} based conditioning scheme. However, we had difficulty achieving sufficient convergence and in justifying the additional compute cost. Nevertheless, related approaches could be a promising area for future study. 

\subsection{Limitations}
We believe our method to be a valuable step towards in-the-wild single-view novel view synthesis but we acknowledge several limitations. While we demonstrate our method to be competitively geometrically consistent, it is not inherently 3D or temporally consistent. Noticeable flicker and other artifacts are sometimes visible in rendered sequences.

While our model generally produces plausible renderings, it may not always perfectly transfer details from the input. On ShapeNet, this sometimes manifests as an inability to replicate the angle of car tires or the style of windows across the line of symmetry; on more complex datasets, the model sometimes struggles to transfer fine details. We use a relatively lightweight, ResNet-backed Deeplab feature encoder. A more powerful encoder, potentially one that makes use of attention to improve long-range information flow, may resolve these issues.

\end{document}